\documentclass{article}
\usepackage[final]{colm2026_conference}
\usepackage[T1]{fontenc}

\usepackage{microtype}
\usepackage{graphicx}
\usepackage{subcaption}
\usepackage{booktabs}
\usepackage{hyperref}
\usepackage{url}
\usepackage{lineno}
\usepackage{amsmath}
\usepackage{amssymb}
\usepackage{mathtools}
\usepackage{amsthm}
\usepackage[capitalize,noabbrev]{cleveref}
\usepackage[textsize=tiny]{todonotes}
\usepackage{array}
\usepackage{colortbl}
\usepackage{xcolor}
\usepackage{multirow}
\usepackage{listings}
\usepackage{float}
\usepackage{titlesec}
\usepackage{enumitem}
\usepackage{xkeyval}
\usepackage[most]{tcolorbox}
\usepackage{svg}
\usepackage{wrapfig}
\usepackage{xspace}
\usepackage{wrapfig}

\theoremstyle{plain}

\theoremstyle{definition}

\theoremstyle{remark}

\definecolor{codegray}{rgb}{0.5,0.5,0.5}
\definecolor{backcolour}{rgb}{0.97,0.97,0.97}
\definecolor{myPink}{HTML}{e73d8c}
\definecolor{darkblue}{rgb}{0, 0, 0.5}
\hypersetup{colorlinks=true, citecolor=darkblue, linkcolor=darkblue, urlcolor=darkblue}

\lstdefinestyle{mystyle}{
    backgroundcolor=\color{backcolour},
    commentstyle=\color{codegray},
    keywordstyle=\color{blue},
    numberstyle=\tiny\color{codegray},
    stringstyle=\color{red},
    basicstyle=\ttfamily\small,
    breakatwhitespace=false,
    breaklines=true,
    captionpos=b,
    keepspaces=true,
    numbers=left,
    numbersep=5pt,
    showspaces=false,
    showstringspaces=false,
    showtabs=false,
    tabsize=4
}
\newcommand{\blue}[1]{\textcolor{blue}{#1}}
\newcommand{\red}[1]{\textcolor{red}{#1}}
\newcommand{\think}{Thinking}
\newcommand{\nothink}{NT}

\newcommand{\bench}{\texttt{LogicIFEval}}
\newcommand{\benchmini}{\texttt{LogicIFEval-mini}}
\newcommand{\framework}{\texttt{LogicIFGen}}
\newcommand{\trainset}{\texttt{LogicIFTrain}}
\newcommand{\task}{\texttt{LogicIF}}

\title{LogicIF: Towards Complex Logic Instruction Following}

\author{\textbf{Mian Zhang}\textsuperscript{1}$^\dagger$, %
\textbf{Shujian Liu}\textsuperscript{5}, %
\textbf{Sixun Dong}\textsuperscript{2}$^\dagger$, %
\textbf{Ming Yin}\textsuperscript{3}$^\dagger$, %
\textbf{Yebowen Hu}\textsuperscript{4}$^\dagger$ \\%
\textbf{Xun Wang}\textsuperscript{5}, %
\textbf{Simin Ma}\textsuperscript{5}, %
\textbf{Song Wang}\textsuperscript{5}, %
\textbf{Sathish Reddy Indurthi}\textsuperscript{5}, %
\textbf{Haoyun Deng}\textsuperscript{5} \\%
\textbf{Zhiyu Zoey Chen}\textsuperscript{1}$^{*}$, %
\textbf{Kaiqiang Song}\textsuperscript{5}$^{*}$ \\
\textsuperscript{1}UTD \qquad
\textsuperscript{2}ASU \qquad
\textsuperscript{3}Duke \qquad
\textsuperscript{4}UCF \qquad
\textsuperscript{5}Zoom
}
\begin{document}

\ifcolmsubmission
\linenumbers
\fi

\maketitle

\renewcommand\thefootnote{} 
\footnotetext{$\dagger$~Work done during internship at Zoom Communications.}   
\footnotetext{$*$ Corresponding author.}
\renewcommand\thefootnote{\arabic{footnote}}

\begin{figure*}[h!]
    \centering
    % First image
    \begin{minipage}[t]{0.34\linewidth}
        \centering
        \vspace{-13.5em}
        \includegraphics[width=\linewidth]{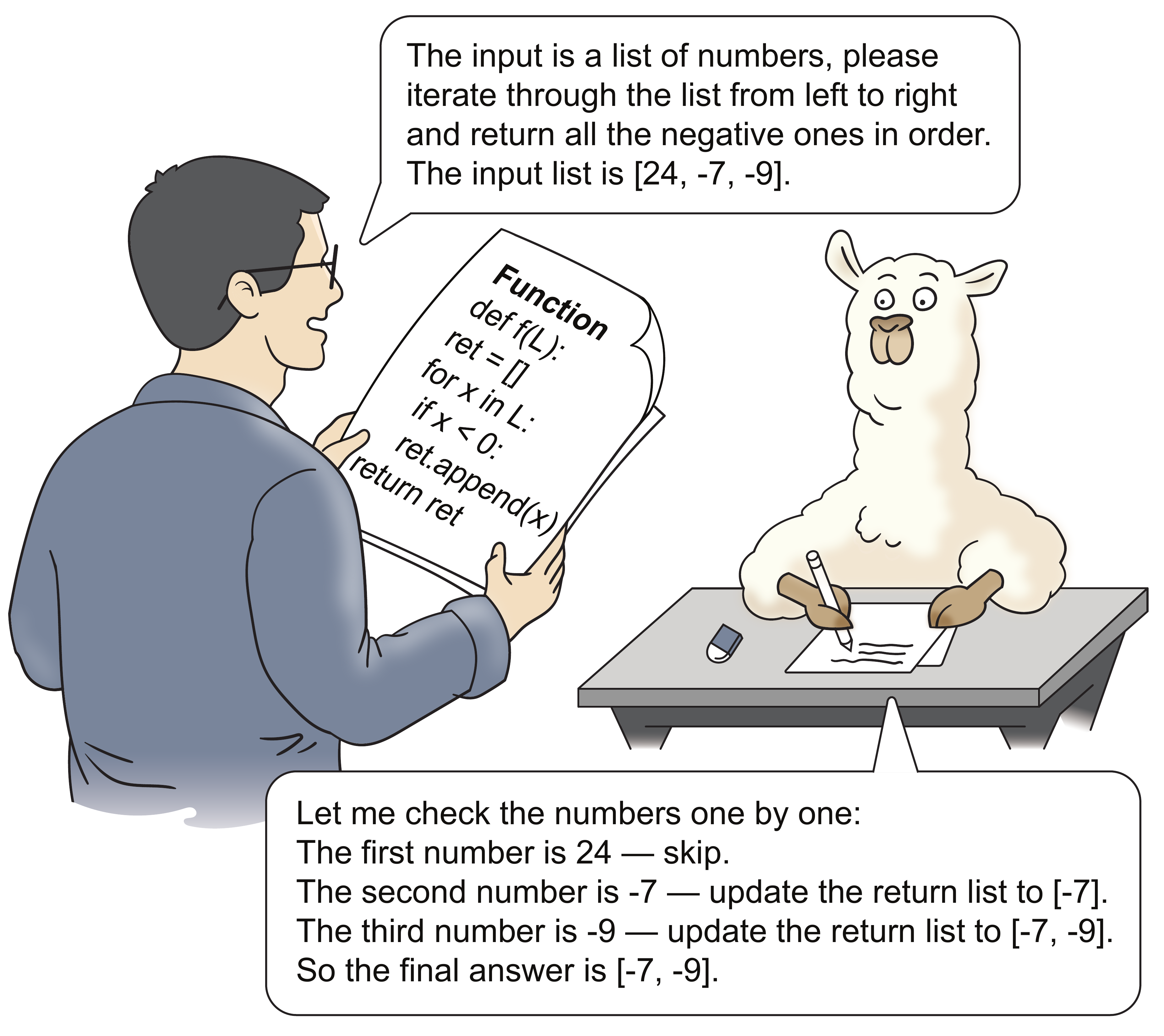}
        \label{fig:introimage}
    \end{minipage}%
    % \hspace{0.02\linewidth} % horizontal space between the figures
    % Second image
    \begin{minipage}[t]{0.54\linewidth}
        \centering
        \includegraphics[width=\linewidth]{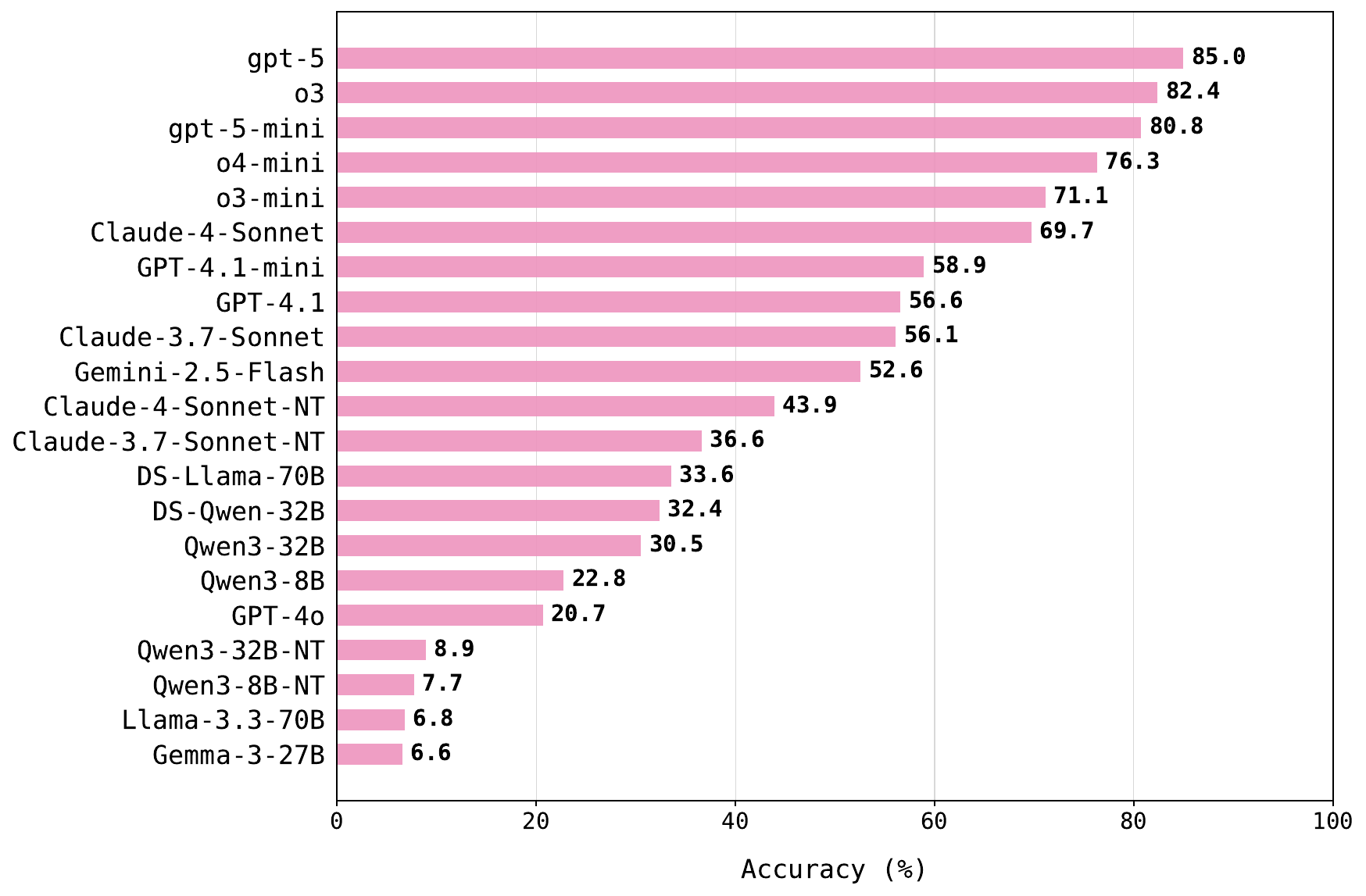}
        \label{fig:overall}
    \end{minipage}
    \caption{\textbf{(Left)} Logic Instruction Following Test: LLMs are required to follow only the natural language instruction to simulate every logic of a code function via generating text. \textbf{(Right)} Overall performance of evaluated models on \bench{}; NT denotes NoThinking.}\label{fig:tiser}
\end{figure*}

\begin{abstract}
Instruction following has catalyzed the recent era of Large Language Models (LLMs) and is the foundational skill underpinning more advanced capabilities such as reasoning and agentic behaviors. As tasks grow more challenging, the logic structures embedded in natural language instructions becomes increasingly intricate. However, how well LLMs perform on such logic-intensive instructions remains under-explored. We propose Logic Instruction Following (\task{}), a task requiring models to precisely follow and simulate every logic step within complex instructions. To generate these instructions, we construct \framework{}, a scalable, automated framework for generating verifiable instructions from code functions, which can naturally express rich logic such as conditions, loops, nesting, and function calls. Using this framework, we curate a collection of complex code functions to construct \bench{}, a benchmark comprising 426 verifiable, human-examined, logic-rich instructions. Our experiments demonstrate that current state-of-the-art LLMs still struggle to correctly follow the instructions in \bench{}. Most LLMs can only follow fewer than 60\% of the instructions, revealing significant deficiencies in their capacity to handle instructions that involve complex logic structures. To further demonstrate the efficacy of our framework, we collect an extensive set of code functions across a wide range of logic difficulties to generate \trainset{}, a fully synthesized training set featuring 27,324 verifiable instructions. Our results show that a model trained on \trainset{} using reinforcement learning significantly outperforms the base model on both in-domain and out-of-domain benchmarks. All the data, codes and models are released at \href{https://github.com/mianzhang/LogicIF}{https://github.com/mianzhang/LogicIF}.

\end{abstract}

\section{Introduction}

Before ChatGPT~\citep{Ouyang2022-ic}, chatbots built on earlier models such as GPT-1~\citep{Radford2018-qp}, GPT-2~\citep{Radford2019-dq}, GPT-3~\citep{Brown2020-vp}, and other architectures struggled to generate coherent and contextually appropriate utterances. At that time, it was difficult to imagine that such models could assist with everyday tasks. With the emergence of instruction-following capabilities, large language models (LLMs) can now accurately understand basic human intentions and even leverage tools to perform a wide range of productivity-enhancing tasks, such as \href{https://openai.com/index/introducing-deep-research}{deep research}, \href{https://cursor.com}{coding assistance}, and scientific discovery~\citep{AI4Science2023-gk}. Instructions for these tasks may contain rich logic structures, including sequencing, loops, nesting, recursion, and backtracking. Previous instruction-following evaluations typically focus on instructions that constrain response format (e.g., “fewer than 300 words”) or content (e.g., “in Shakespeare’s tone”)~\citep{Zhou2023-bp,Qin2024-cs,Jiang2024-hu}, and seldom examine \textit{how well LLMs follow instructions with rich logic structures}.

To address this gap, we propose \textbf{Logic} \textbf{I}nstruction \textbf{F}ollowing (\task{}), a task that requires models to precisely follow and simulate every logic step in complex instructions. This setting differs from traditional instruction following, which primarily emphasizes constraint satisfaction. An instruction in \task{} consists of an ordered sequence of sub-steps $\mathcal{S} = (s_1, s_2, \dots, s_k)$; models are expected to move from an initial state to a final state by strictly following these steps. In real-world scenarios, logic following and constraint fulfillment can coexist and entangle. To create instructions with intensive logic for studying \task{} (independent of constraint fulfillment), we develop \framework{}, a \textbf{scalable}, \textbf{automated} framework that generates \textbf{verifiable} instructions from \textbf{code functions}, which naturally contain rich logic structures. Given specific test inputs, models are expected to rely \textbf{solely} on the natural-language instruction to simulate the logic of the underlying code function and produce the corresponding outputs, analogous to being verbally guided by an examiner to process the input data step-by-step (see Fig.~\ref{fig:tiser} (Left)). Models are prohibited from writing code or using external tools; instead, they must execute the logic through text generation alone. This setting aligns closely with instruction-following evaluation, as most tasks require a model to explicitly unfold underlying logic. For example, when given an instruction such as “repeat asking the user for clarification until you fully understand the user's intent,” the model must perform each logic step through natural language alone. \framework{} obtains reference labels by executing the code function on the test inputs. By comparing model outputs with these reference labels, we can directly verify whether a model follows the natural-language instruction correctly. In addition, \framework{} incorporates state trackers to monitor intermediate logic flow, enabling us to double-check whether models faithfully adhere to an instruction’s internal logic rather than hallucinating final results.

Second, we construct a benchmark called \bench{} using \framework{}, which contains 426 verifiable, logic-rich instructions paired with test cases. The functions used to generate \bench{} are solutions to challenging simulation problems from competitive programming platforms, \href{https://codeforces.com}{CodeForces} and \href{http://poj.org}{POJ}. These solutions are particularly suitable for instruction-following evaluation because they require models to faithfully emulate complex, step-by-step processes and state transitions, often involving intricate control flow, edge-case handling, and coordination among multiple logic components.

Experimental results show that most popular LLMs correctly follow fewer than 60\% of the instructions in \bench{}, revealing a substantial deficiency in their instruction-following ability (see Fig.~\ref{fig:tiser} (Right)). Open-source models continue to lag behind frontier models such as the OpenAI o-series and Anthropic Claude. As logic complexity increases, models find it increasingly difficult to accurately interpret and follow instructions. We also observe that incorporating explicit thinking before response generation can improve instruction-following performance. Further error analysis and case studies reveal key failure modes and highlight promising directions for improving LLMs’ ability to follow logic-rich instructions.

% We will release \framework{} and \bench{}, as well as a compute-friendly version of the benchmark, \benchmini{}

To demonstrate the practical utility of our framework, we leverage \framework{} to curate \trainset{}, a fully synthesized training set featuring 27,324 verifiable instructions. We employ Reinforcement Learning from Verifiable Rewards (RLVR) to fine-tune a Qwen3-1.7B base model, and our experimental results indicate that the trained model achieves substantial performance gains of +16.7 points on \bench{}. More importantly, this specialized training exhibits strong generalization to diverse out-of-domain tasks, including logic reasoning (ZebraLogic: +31.4), coding (HumanEval: +4.6), mathematical reasoning (MATH-500: +4.3), and graduate-level general knowledge (GPQA-Diamond: +3.1). These findings suggest that the ability to precisely follow logic-rich instructions serves as a foundational skill that bolsters a wide range of advanced cognitive capabilities.

% \github{} \textcolor{myPink}{https://github.com/mianzhang/LogicIF}.

\section{\framework{}: Verifiable Instruction Generation from Code}
\begin{figure*}[t]
    \centering
    \includegraphics[width=0.95\linewidth]{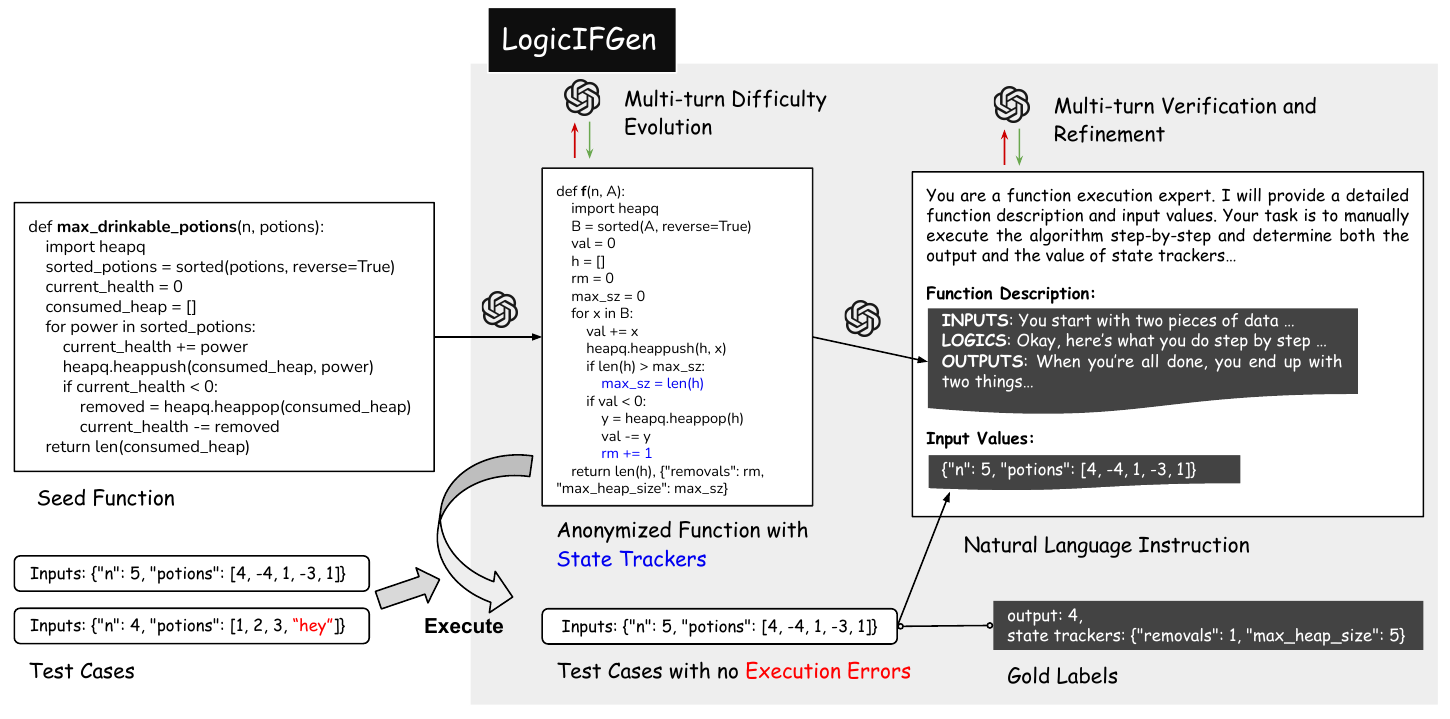}
    \caption{Pipeline of \framework{}. Given a seed function and its corresponding test cases, \framework{} generates natural language instructions along with gold labels, which include both the function outputs and the values of state trackers. 1) The input function is first anonymized and augmented with \textcolor{blue}{state trackers}. 2) The anonymized function is then translated into a natural language description, producing a instruction that precisely describes its logic and expected behavior with test cases verified to have no \textcolor{red}{execution errors}. 3) Finally, the test cases are executed on the anonymized function to obtain the gold labels.}
    \label{fig:logicifgen}
\end{figure*}

We introduce \framework{}, a framework that synthesizes verifiable, logic-rich instructions from \textbf{code functions}. Each instruction provides a comprehensive, step-by-step natural language description of the function’s behavior, clearly specifying input/output formats and detailing all relevant control flows and data processing steps. Models are then expected to follow these instructions \textbf{without access to the source code} to process inputs and reproduce the function's outputs. Fig.~\ref{fig:logicifgen} illustrates our instruction generation framework with a running example. Additionally, \framework{} integrates \textit{Multi-turn Difficulty Evolution} to dynamically adjust instruction complexity and \textit{Multi-turn Verification and Refinement} to verify and refine each instruction to accurately captures the function's logic.

% fully and accurately captures the function's logic. 
% Details are provided below.

\paragraph{Anonymized Function with State Trackers}
The first step in \framework{} is to anonymize the original function so that \textbf{only its data operations and control logic remains}. \framework{} use an LLM to replace the function and variable names with simple, generic identifiers, single characters or common placeholders, thereby stripping away any semantic hints that could enable models to leverage domain-specific knowledge. Moreover, \framework{} asks an LLM to augment the function with \textbf{state trackers}, which are variables designed to log specific runtime states of the function, reflecting the function’s internal logic and execution flow. These include, for example, the number of iterations of a \texttt{for} loop, the number of times an \texttt{if} block is executed, or the maximum length reached by a dynamically list. Models instructed via natural language are required to reproduce not only the function’s outputs but also the values of state trackers. We consider the instruction to be correctly followed only when both the final output and the state tracker values match the ground truth. This ensures that models truly comprehend and execute the intended logic, rather than relying on shortcuts or hallucinating plausible outputs. The prompt used for this step is shown in Fig.~\ref{fig:prompt_add_stat}. Furthermore, to increase the logic complexity of the function, the \textit{Multi-turn Difficulty Evolution} module in \framework{} prompts an LLM to evolve the code function by introducing more diverse compositions of logic units, such as loops, function calls, and recursion (see Appendix Fig.~\ref{fig:prompt_evolve} for the full prompt).

% This anonymization is crucial for fairly evaluating a model’s ability to follow the generated instructions purely based on the described logic.

\paragraph{Natural Instruction Generation} 
\framework{} uses an LLM (see Appendix Fig.~\ref{fig:prompt_gen_desc} for the prompt) to generate detailed natural language instructions based on the anonymized function with state trackers. The instructions adopt a conversational style (e.g., “Now you need to…”, “Next, go through…”) to guide step-by-step execution. Each instruction is crafted to be sufficiently precise so that an LLM can follow it to conduct all the data operations without access to the source code, producing identical outputs and state tracker values. To ensure the instructions fully and correctly capture the sub-tasks and logic of the function, \framework{} incorporates the \textit{Multi-turn Verification and Refinement} module. In this module, an LLM (see Appendix Fig.~\ref{fig:prompt_verify_desc} for the prompt) reviews the generated instructions, checking for comprehensive coverage of all code operations and flagging potential omissions, such as loop conditions, variable updates, or edge case handling. The instructions are iteratively refined based on this feedback until they accurately reflect the complete operations and logic of the function. The full function description of the running example in Fig.~\ref{fig:logicifgen} is in Appendix~\ref{apx:func_desc}.

\paragraph{Test Case Filtering and Gold Result Generation}
At this stage, we have two modalities representing the same operations and logic: the anonymized code function and the corresponding natural language instruction. Given the same input data, LLMs are expected to follow the instruction and produce results identical to those generated by the code function, including both the function outputs and the values of the state trackers. \framework{} generates these gold labels by executing the test cases on the anonymized function. Prior to this, it performs a preliminary execution step to filter out test cases that may trigger execution errors due to modifications introduced to function during the multi-turn difficulty evolution process. This ensures that only valid and executable test cases are retained.
% for final gold label generation.

\paragraph{Quantifiable Instruction Complexity}
One of the advantages of \framework{} is that the difficulty of the generated instructions is \textbf{quantifiable}: the anonymized function and natural language instruction describe the same operations and logic so we could use the code function as a proxy to analyze the difficulty of the instruction. We use Python's \href{https://docs.python.org/3/library/ast.html}{Abstract Syntax Tree (AST) package} to traverse the syntactic structure of each function and calculates several key measures. \textit{Cyclomatic Complexity} ($C$) quantifies the total number of control flow decision points (such as if/elif statements, for/while loops, and try/except blocks), representing the number of linearly independent execution paths through the code. \textit{Nesting depth} ($D$) tracks the maximum depth of nested control structures. \textit{Function Call Count} ($F$) counts the number of function invocation, including both built-in and user-defined functions. \textit{Function length} ($L$) is calculated as the number of lines spanned by the function definition. We adopt an intuitive weighting scheme for aggregating these AST-based measures based on their cognitive complexity, which has been proven effective in measure the logic complexity of codes~\citep{Munoz-Baron2020-fz}:
\begin{equation}\label{eq:score}
\text{Score} = D \times 3 + F \times 2 + C \times 1 + L \times 0.5.    
\end{equation}
Experiments in Section~\ref{sec:performance} also show the effectiveness of the intuitive weights. Weighting methods like learning-based methods~\citep{Sepidband2025-vw} could be explored in future. 
 
\section{\bench{}: Benchmark for Logic Instruction Following}
In the previous section, we introduce \framework{}, which generates instructions and gold labels from code functions. However, we observed that the complexity of the seed function is crucial for producing challenging instructions. To this end, we curate a collection of simulation problem solutions and corresponding test cases from competitive programming platforms, including Codeforces and POJ. Specifically, for Codeforces, we select problems tagged with “implementation” and a difficulty score above 1700. For POJ, we include difficult simulation problems as identified by various online users. These functions require models to faithfully emulate complex, step-by-step processes and state transitions, involving intricate control flow, edge case handling, and the coordination of multiple logic elements, making them ideal sources for generating complex instructions used to evaluate the instruction-following ability. Then we construct \bench{} based on these functions.

\paragraph{Data Filtering} We use a two-stage filtering process to make the seed functions more diverse and remove test cases not suitable for the instruction-following evaluation: \textit{1) Seed Function Filtering}: We remove duplicate or highly similar functions to avoid redundancy. Specifically, we use OpenAI's \href{https://platform.openai.com/docs/models/text-embedding-3-small}{text-embedding-3-small} model to compute embeddings of each function and calculate pairwise cosine similarities. If the similarity between two functions is greater than 0.7, we consider them near-duplicates. For each such pair, we keep only the longer function (measured by code length), since longer implementations often include richer logic and contribute to more challenging instruction-following tasks. \textit{2) Test Case Filtering}: We remove test cases with unusual or problematic outputs to keep the dataset clean and manually executable. Specifically, we discard cases where: I) The values of state trackers are greater than or equal to 50 (to avoid overly large internal states). II) The output values have excessive precision (more than six decimal places). III) The state tracker dictionaries are malformed or incorrectly formatted. IV) The input values are too large (magnitude exceeding $10^7$). After this step, we remove functions with fewer than 3 test cases to ensure that each function has enough test coverage for reliable evaluation. These filtering steps are designed to prevent failures caused by large loops, deep recursion, or complex numerical computations that could overwhelm model capacity. \textbf{This ensures that any model failures are attributable to weaknesses in instruction following, rather than the inherent difficulty of execution}. Originally, we collected 1,107 seed functions. After these filtering steps, 426 unique functions and 3,050 test cases remain.
% The final dataset is easy to understand and execute step-by-step, staying consistent with our goal of generating clear and verifiable instructions. 

\paragraph{Data Generation and Human Verification}
We apply \framework{} using o4-mini as the generation LLM on the filtered seed functions and test cases, resulting in \bench{}, a benchmark comprising 426 complex instruction-following tasks. The numbers of turns used for Multi-turn Difficulty Evolution and Multi-turn Verification and Refinement are 1 and 3, respectively. Instructions that still fail verification after 3 turns are discarded. To evaluate the quality of the generated instructions, we hired five PhD-level experts in computer science (our co-authors) to conduct manual verification. The annotators were instructed to verify whether each line of the anonymized function is accurately described in the corresponding natural language instruction. Each instruction is examined by two independent experts. According to their assessment, 97\% of the instructions fully and correctly capture the underlying function logic and their agreement is 97.79\%, demonstrating the effectiveness of \framework{} in transforming code functions into natural language instructions. More details human evaluation process could be found in Appendix~\ref{sec:human_verify}. We further categorize the functions into difficulty levels based on the complexity scores computed by Equation~\ref{eq:score}. The final benchmark contains 142 easy, 145 medium, and 139 hard instructions.

% using tercile-based thresholds derived from the

\definecolor{sectiongray}{gray}{0.90}
\definecolor{headergray}{gray}{0.96}

\begin{table*}[t]
  \centering
  \renewcommand{\arraystretch}{1.1}
    \caption{Question-level model performance (\%) by complexity. “Output (State)” denotes a question is correct if the model produces the correct outputs (state trackers) for all associated test cases. “Both” denotes both the output and state trackers match. Overall performance is highlighted in \colorbox{pink!50}{pink}. Performance of each difficulty level is in \colorbox{blue!20}{blue}. The “Average” refers to micro-averaging, which is computed by summing the number of solved questions across the difficulty levels and dividing by the total number of questions.}
  \label{tab:model_performance_by_complexity}
  \resizebox{0.95\textwidth}{!}{%
  \begin{tabular}{@{}l cc>{\columncolor{blue!20}}c cc>{\columncolor{blue!20}}c cc>{\columncolor{blue!20}}c cc>{\columncolor{pink!50}}c@{}}
    \toprule
    \multicolumn{1}{c}{} &
      \multicolumn{3}{c}{\textbf{Easy (142)}} &
      \multicolumn{3}{c}{\textbf{Medium (145)}} &
      \multicolumn{3}{c}{\textbf{Hard (139)}} &
      \multicolumn{3}{c}{\textbf{Average}} \\
    \textbf{Model} 
      & Output & State & \textbf{Both}
      & Output & State & \textbf{Both}
      & Output & State & \textbf{Both}
      & Output & State & \textbf{Both} \\
    \cmidrule(lr){2-4} \cmidrule(lr){5-7} \cmidrule(lr){8-10} \cmidrule(lr){11-13}
    \rowcolor{sectiongray}
    \multicolumn{13}{l}{\textbf{\think{} Models}} \\
    gpt-5 & 94.37 & 95.07 & 90.85 & 97.24 & 91.03 & 89.66 & 88.49 & 82.01 & 74.10 & \textbf{93.43} & \textbf{89.44} & \textbf{84.98} \\
    o3 & 94.37 & 90.14 & 89.44 & 93.10 & 87.59 & 84.83 & 84.89 & 79.86 & 72.66 & 90.85 & 85.92 & 82.39 \\
    gpt-5-mini & 91.55 & 92.96 & 88.73 & 93.79 & 84.83 & 82.07 & 86.33 & 78.42 & 71.22 & 90.61 & 85.45 & 80.75 \\
    o4-mini & 93.66 & 90.14 & 87.32 & 91.72 & 81.38 & 77.93 & 81.29 & 68.35 & 63.31 & 88.97 & 80.05 & 76.29 \\
    o3-mini & 89.44 & 87.32 & 83.10 & 89.66 & 78.62 & 73.79 & 69.06 & 61.87 & 56.12 & 82.86 & 76.06 & 71.13 \\
    Claude-4-Sonnet & 91.55 & 87.32 & 81.69 & 96.55 & 77.93 & 75.86 & 73.38 & 58.27 & 51.08 & 87.32 & 74.65 & 69.72 \\
    \midrule
    \multicolumn{13}{c}{\textit{--- Models below this line have Average Accuracy $<$ 60\% ---}} \\
    \midrule
    Claude-3.7-Sonnet & 81.69 & 76.76 & 70.42 & 79.31 & 64.14 & 57.93 & 57.55 & 45.32 & 39.57 & 73.00 & 62.21 & 56.10 \\
    Gemini-2.5-Flash & 79.58 & 75.35 & 72.54 & 64.14 & 55.17 & 53.10 & 41.01 & 34.53 & 31.65 & 61.74 & 55.16 & 52.58 \\
    DS-Qwen-32B & 66.20 & 59.86 & 50.70 & 57.93 & 36.55 & 30.34 & 38.13 & 22.30 & 15.83 & 54.23 & 39.67 & 32.39 \\
    DS-Llama-70B & 69.01 & 66.20 & 54.93 & 53.10 & 36.55 & 29.66 & 38.13 & 24.46 & 15.83 & 53.52 & 42.49 & 33.57 \\
    Qwen3-32B & 71.13 & 57.75 & 50.00  & 50.34 & 32.41 & 27.59 & 34.53 & 20.86 & 13.67 & 52.11 & 37.09 & 30.52 \\
    
    Qwen3-8B & 52.11 & 55.63 & 40.85 & 40.00 & 27.59 & 20.00 & 25.18 & 15.83 & 7.19  & 39.20 & 33.10 & 22.77 \\
    \addlinespace[1ex]
    \rowcolor{sectiongray}
    \multicolumn{13}{l}{\textbf{NoThinking Models}} \\
    GPT-4.1-mini & 85.92 & 81.69 & 74.65 & 82.07 & 64.83 & 60.00 & 71.22 & 49.64 & 41.73 & 79.81 & 65.49 & 58.92 \\
    GPT-4.1 & 83.10 & 78.87 & 71.13 & 84.83 & 64.83 & 58.62 & 64.75 & 51.80 & 39.57 & 77.70 & 65.26 & 56.57 \\
    Claude-4-Sonnet-NT & 80.28 & 76.06 & 69.01 & 63.45 & 46.21 & 37.93 & 44.60 & 30.22 & 24.46 & 62.91 & 50.94 & 43.90 \\
    Claude-3.7-Sonnet-NT & 73.24 & 66.20 & 57.04 & 61.38 & 42.07 & 34.48 & 43.88 & 24.46 & 17.99 & 59.62 & 44.37 & 36.62 \\
    GPT-4o & 59.86 & 49.30 & 38.73 & 32.41 & 24.14 & 17.93 & 12.95 & 10.07 & 5.04 & 35.21 & 27.93 & 20.66 \\
    \midrule
    \multicolumn{13}{c}{\textit{--- Models below this line have Average Accuracy $<$ 10\% ---}} \\
    \midrule
    Qwen3-32B-NT & 33.10 & 29.58 & 19.72 & 18.62 & 9.66 & 4.14 & 9.35 & 4.32 & 2.88 & 20.42 & 14.55 & 8.92 \\
    Llama-3.3-70B & 30.28 & 25.35 & 15.49 & 10.34 & 4.83 & 2.76 & 7.91 & 3.60 & 2.16 & 16.20 & 11.27 & 6.81 \\
    Gemma-3-27B & 28.17 & 26.76 & 15.49 & 10.34 & 5.52 & 2.07 & 5.76 & 4.32 & 2.16 & 14.79 & 12.21 & 6.57 \\
    Qwen3-8B-NT & 28.17 & 29.58 & 16.90 & 14.48 & 8.28 & 4.83 & 5.04 & 5.04 & 1.44 & 15.96 & 14.32 & 7.75 \\
    \bottomrule
  \end{tabular}
  }
\end{table*}

\paragraph{Benchmark Statistics and Release}
The instructions in \bench{} have average of 3,428 characters and 662 words. The functions used to generate instructions have an average cyclomatic complexity of 11.10 and a maximum nesting depth of 3.16. In total, the benchmark includes 2,049 loops, 2,253 conditional statements, and 5,289 function calls, reflecting diverse control flow patterns. Each function is evaluated with an average of 7.2 test cases, resulting in 3,050 test cases overall. These statistics highlight the benchmark's scale and its focus on challenging instruction-following in LLMs. To support researchers with limited computational resources, we also release a representative mini-benchmark, \benchmini{}, which consists of 102 functions sampled in a stratified manner based on the complexity scores. Our experimental results in Section~\ref{sec:performance} show that constructing the min-benchmark based on the complexity scores is effective.

\section{Model Performance on \bench{}}\label{sec:performance}

We test 21 frontier LLMs on \bench{}. \textbf{1) \think{} Models}: Models incorporating explicit thinking process before generating a response, including gpt-5, gpt-5-mini, o3, o4-mini, o3-mini,  Claude-3.7-Sonnet and Claude-4-Sonnet, Gemini-2.5-Flash, Qwen3-32B and Qwen3-8B, DeepSeek-R1-Distill(DS)-Llama-70B and DS-Qwen-32B. \textbf{2) NoThinking (\nothink{}) Models}: Models directly give a response without explicit thinking: GPT-4.1, GPT-4.1-mini, GPT-4o, Claude-3.7-Sonnet-\nothink{} and Claude-4-Sonnet-\nothink{}, Gemma-3-27B, Qwen3-32B-\nothink{} and Qwen3-8B-\nothink{}, and Llama-3.3-70B(-Instruct).
% \begin{itemize}[leftmargin=3em]
% \end{itemize}

\paragraph{Inference Setting} For closed-source models, we use the default temperature setting provided by the respective API. For open-source models, we adopt the official recommended inference settings; for example, for the Qwen3 series reasoning models, the temperature is set to 0.6. If no specific recommendation is available, we use a temperature of 1.0 by default. For all models, we set the maximum number of generated tokens to 16k to ensure that the models have sufficient token budgets to follow instructions and process the inputs.

\paragraph{Main Results} 
A model is considered to successfully follow a natural language instruction if it passes all associated test cases by producing both the correct outputs and accurate state tracker values. The last column of Table~\ref{tab:model_performance_by_complexity} report the overall accuracies across evaluated models. \textbf{The top-performing models are the OpenAI gpt-5, o-series and Claude-4-Sonnet}, with the best-performing model, gpt-5, achieving an accuracy of 84.98\%. These results highlight the strong instruction-following and logic execution capabilities of advanced proprietary LLMs. In contrast, most of other models can only have less than 60\% accuracies and there is huge room for improvement. For example, GPT-4o achieves only 20.66\%, significantly underperforming relative to other OpenAI models. \textbf{Besides, widely used open-source models still lag significantly behind.} For example, Qwen3-32B-NT, Gemma-3-27B, and Llama-3.3-70B all score below 10\%, failing to correctly execute the multi-step logic and state tracking required by \bench{}, which highlights a clear performance gap between commercial and open-source LLMs on this benchmark. We also notice that \textbf{explicit thinking before response can potentially improve instruction following for LLMs.} For instance, Claude-4-Sonnet achieves 69.72\% accuracy, notably outperforming Claude-4-Sonnet-\nothink{} (43.9\%) and Qwen3-32B (30.52\%) outperforms Qwen3-32B-\nothink{} (8.92\%). Similarly, OpenAI's thinking models, gpt-5 and o-series, perform substantially better than other non-thinking models from OpenAI. We also report model performance on \benchmini{} in Fig.~\ref{fig:overall_mini}. The results show that models exhibit nearly identical rankings compared to the full benchmark, indicating that \benchmini{} effectively represents the whole benchmark and preserves a similar distribution.

\paragraph{Results Across Difficulty Levels} 
Table~\ref{tab:model_performance_by_complexity} presents model performance across three difficulty levels in \bench{}, with separate evaluations for output correctness, state tracker correctness, and their intersection (“Both”). We could see that \textbf{all models show a clear degradation in performance as difficulty increases} (see \colorbox{blue!20}{blue columns}), validating the effectiveness of our AST-based complexity scoring strategy. For instance, gpt-5 drops from 90.85\% on Easy tasks to 74.10\% on Hard ones, and GPT-4.1-mini from 82.07\% to 41.73\%. This trend confirms that our benchmark's stratification meaningfully reflects logic complexity of instructions. \textbf{Second, output accuracy consistently exceeds state tracker accuracy across nearly all models}, especially as complexity increases. For example, GPT-4.1-mini achieves 71.22\% output accuracy on Hard instructions, but only 49.64\% in state tracking. This implies that models may generate correct answers without strictly adhering to the intended logic steps or may follow alternative logic paths. These observations underscore the importance of adding state trackers to supervise the logic flow in complex instruction-following tasks.

\subsection{Analysis}\label{sec:ans}
\begin{wrapfigure}{r}{0.5\textwidth}
    \centering
    \includegraphics[width=\linewidth]{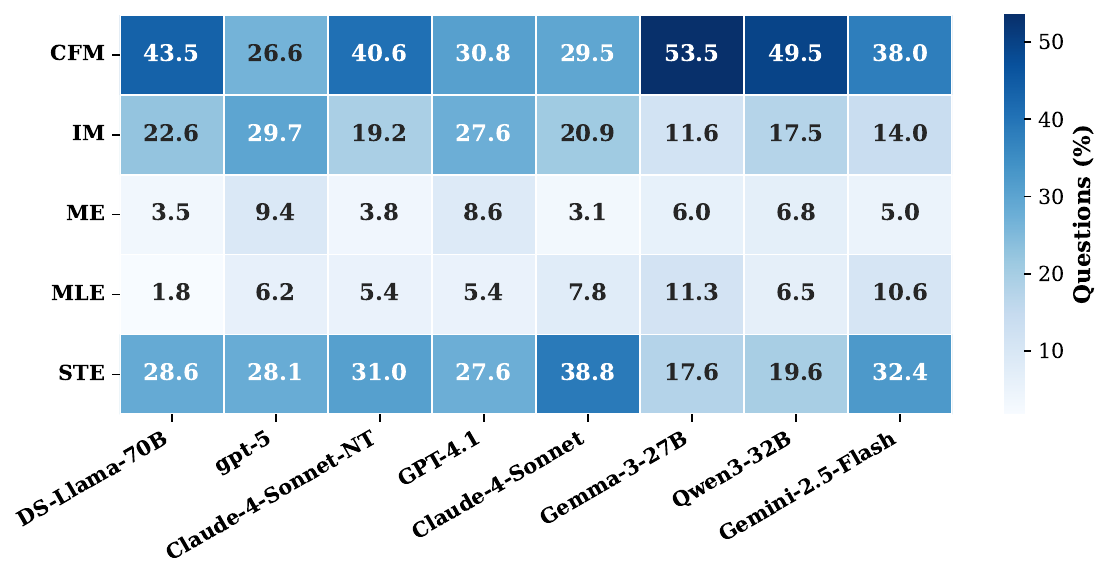}
    \caption{Error Type Distribution in Test Cases}
    \label{fig:err_dist}
\end{wrapfigure}

\paragraph{Failure Modes} 
To understand the reasons of why models can not pass the test cases, we first go through some errors and summarize them as following five types: 
\textbf{1) Control Flow Misexecution} (CFM): incorrect or incomplete execution of core control structures, including wrong iteration counts, improper branching, or mishandled recursion/returns. 
\textbf{2) State Tracking Errors} (STE): failure to correctly maintain or update internal variables or data structures, such as counters, flags, arrays, stacks, or accumulated values. 
\textbf{3) Missing Logic Elements} (MLE): omission of required components (e.g., loops, branches, edge case handling or initialization). 
\textbf{4) Misordered Execution} (ME): performing steps in the wrong sequence, such as using uninitialized variables, premature function calls, or out-of-order updates. 
\textbf{5) Instruction Misinterpretation} (IM): misunderstanding the instruction’s intent, leading to hallucinated steps, misapplied patterns, or ignored constraints. We classify error cases with the help of GPT-4.1.

Fig.~\ref{fig:err_dist} summarizes the error types made by some popular models. We could see that the most frequent error categories across models are Control Flow Misexecution, Instruction Misinterpretation, and State Tracking Errors. In contrast, Missing Logic Elements and Misordered Execution are consistently low (mostly under 10\%), suggesting that \textbf{most models can identify the required logic components (“what to do”) and their approximate ordering (“when to do it”). However, they often struggle with actually executing the logic elements correctly}, either by mismanaging control structures (e.g., loop iterations, function calls), hallucinating or misinterpreting instruction details, or failing to track internal state variables accurately over time. In addition, \textbf{open-source models like Qwen3-32B, DS-Llama-70B, and Gemma-3-27B exhibit especially high rates of Control Flow Misexecution}, up to 53.5\% in the case of Gemma-3-27B. This highlights their difficulty in faithfully reproducing the logic-heavy instruction steps, which require consistent handling of nesting, conditionals, and function boundaries. Besides, \textbf{State Tracking Errors is also a major issue across nearly all models}. For instance, Claude-4-Sonnet and Gemini-2.5-Flash show 38.8\% and 32.4\% error rates respectively, indicating frequent failures in maintaining correct variable states when following the instructions. These include losing track of counters, failing to propagate updates through data structures, or resetting intermediate results incorrectly. This reinforces the importance of evaluating beyond output correctness. We show representative cases for each error type in Fig.~\ref{fig:errors} and the full examples in Appendix~\ref{error_case}. It should be noted that the error types are not strictly mutually exclusive. For example, the missing logic elements error observed in the fourth example of Fig.~\ref{fig:errors} can also be attributed to the incorrect execution of the while condition. In Appendix Section~\ref{sec:error_vs_logic}, we present a fine-grained analysis of how the error rates of different failure modes vary with the logic complexity of the instructions. The results show a clear positive correlation between logic complexity and error frequency: as complexity increases, models produce substantially more errors.

\begin{figure*}[t]
    \centering
    \includegraphics[width=0.9\linewidth]{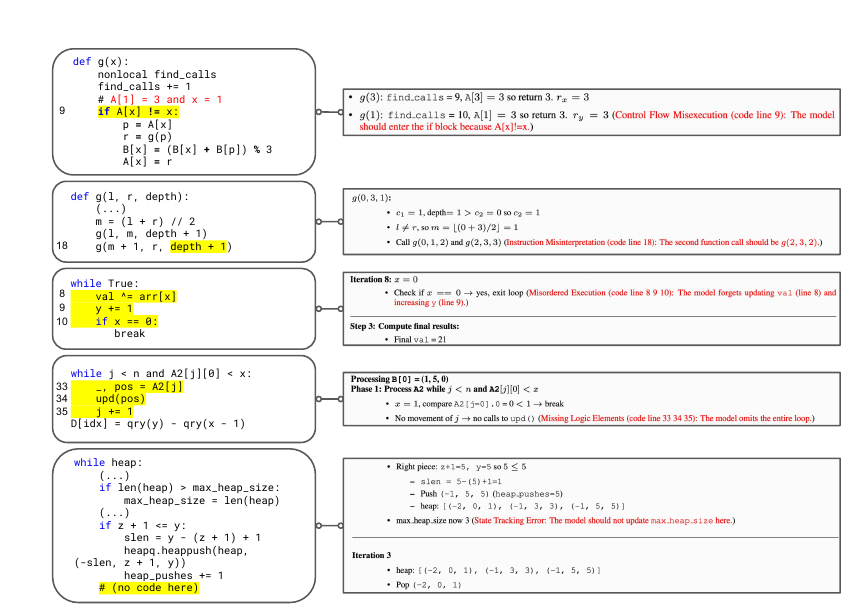}
    \caption{Error Cases: On the left are the \textbf{excerpts} from function codes where the model makes errors. On the right are \textbf{excerpts} from the LLMs’ responses, highlighting their failures across different modes. The explanations for model failures are indicated in \red{red}, and the corresponding code lines are \colorbox{yellow}{highlighted}. Please note that the model only has access to the natural language instruction, which could correctly describe the logic, when solving the tasks; the code is provided here solely to facilitate understanding of the errors.}
    \label{fig:errors}
\end{figure*}

\paragraph{Why Thinking Helps Large LLMs} As indicated in Section~\ref{sec:performance}, incorporating explicit thinking before generating a response can potentially enhance instruction-following performance for large LLMs. To gain deeper insights into this effect, we conduct a case analysis focusing on instances where Claude-4-Sonnet produces correct results, whereas Claude-4-Sonnet-\nothink{} fails. These cases can be categorized into two types: 1) the model arrives at the correct answer during the thinking process itself, and 2) the model articulates a detailed, step-by-step plan to solve the task. Both scenarios suggest that explicit thinking encourages the model to slow down and solve the problem more deliberately, rather than relying on pattern matching or intuitive leaps. We provide two representative examples in the Appendix~\ref{thinking_case}.

% However, in some cases where the model arrives at the correct answer during the thinking process, the model repeats the content of thinking after thinking, reHow to  
\section{Training LLMs to Follow Complex Logic}
In this section, we collect a larger set of data with \framework{} and conduct reinforment learning to answer the following questions:
\begin{itemize}
    \item \textit{Can \framework{} generate useful training data for logic instruction following?}
    \item \textit{Does logic instruction following transfer to other abilities?}
\end{itemize}
\paragraph{Training Data} We utilize the \texttt{code\_contests} dataset \citep{Li2022-vv} as our primary data source, extracting problem descriptions alongside their corresponding Python solutions and test cases. To further enhance diversity, we leverage GPT-5 to generate alternative solutions and supplementary test cases for each problem. These solution functions and test cases are subsequently processed through \framework{} to generate our training set, \trainset{}. While \bench{} focuses on highly complex instructions, \trainset{} comprises 27,324 instructions spanning a broad spectrum of logic complexity, as illustrated in Fig.~\ref{fig:data_dist}.

% \begin{wrapfigure}{r}{0.5\textwidth}
%     \centering
%     \includegraphics[width=\linewidth]{data/qwen3_model_comparison.pdf}
%     \caption{Model Performance Comparison}
%     \label{fig:metric}
% \end{wrapfigure}

\begin{figure*}[h!]
    \centering
    \begin{subfigure}[t]{0.43\linewidth}
        \centering
        % \vspace{-14.5em}
        \includegraphics[width=\linewidth]{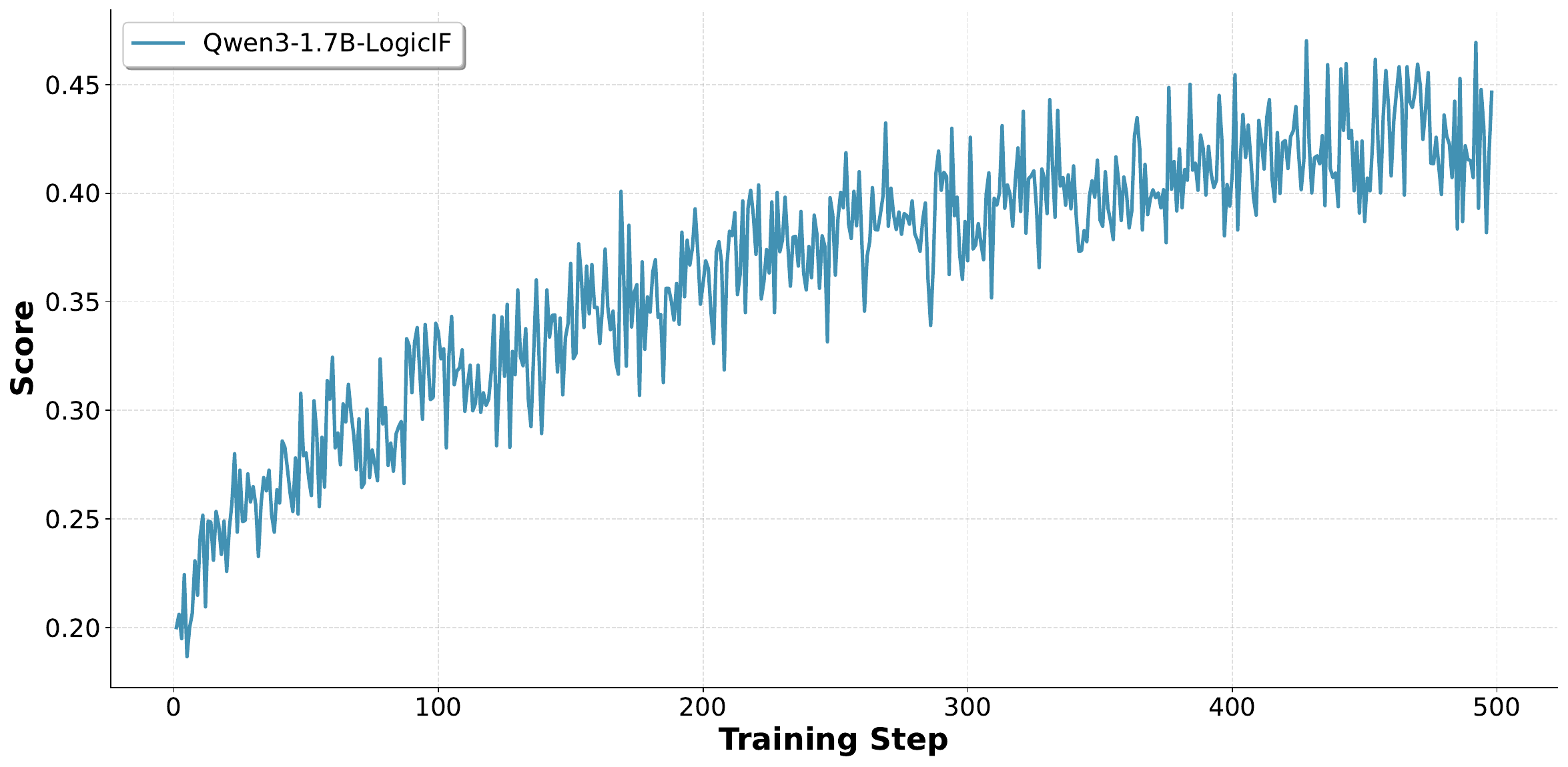}
        \caption{Curve of Training Reward}
        \label{fig:critic}
    \end{subfigure}%
    \hspace{0.01\linewidth}
    \begin{subfigure}[t]{0.55\linewidth}
        \centering
        \includegraphics[width=\linewidth]{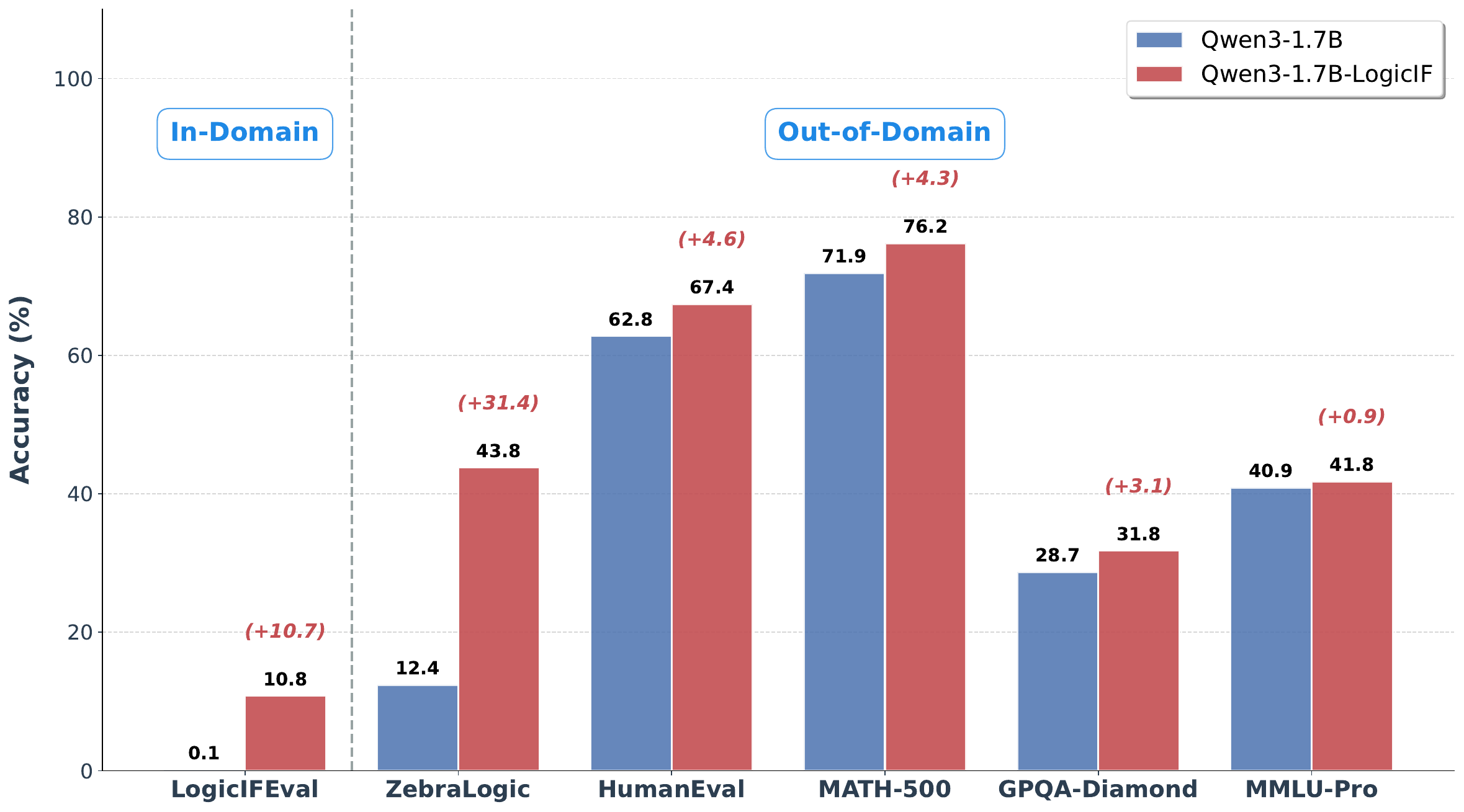}
        \caption{Model Performance Comparison}
        \label{fig:metric}
    \end{subfigure}
    \caption{Training Rewards and Model Performance on Six Benchmarks}\label{fig:train_stat}
\end{figure*}

\paragraph{Training Settings} We employ GRPO \citep{Shao2024-dp} to train the instruct version of Qwen3-1.7B on \trainset{}. The model receives a reward of 1 only if both the final output and the values of the state trackers are correct; otherwise, it receives 0. For each prompt, we set a response group size of 8, utilizing a learning rate of 1e-6 and a total batch size of 512. The training was conducted on four NVIDIA A100 GPUs and completed in 500 steps. We call the trained model as Qwen3-1.7B-LogicIF.

\paragraph{Evaluation} We evaluate our models across six benchmarks, comprising an in-domain benchmark (the mini version of \bench{}) and five out-of-domain benchmarks. The latter cover mathematical reasoning (MATH-500~\citep{Lightman2023-ip}), logic reasoning (ZebraLogic~\citep{lin2025zebralogic}), general knowledge (GPQA-Diamond~\citep{rein2024gpqa}, MMLU-Pro~\citep{wang2024mmlu}), and coding (HumanEval~\citep{chen2021evaluating}). For ZebraLogic, we use greedy decoding to align with the official evaluation protocol. For all other benchmarks, we follow the recommendation from Qwen and use a temperature of 0.7.

\paragraph{Model Comparison and Analysis} The experimental results illustrated in Fig.~\ref{fig:train_stat} demonstrate the effectiveness of training LLMs on logic-rich instructions using the \trainset{}. As shown in Fig.~\ref{fig:critic}, the training reward curve for the Qwen3-1.7B-LogicIF exhibits a steady and consistent upward trend over 500 training steps, indicating that \textbf{\trainset{} provides effective train signals for the policy model to optimize the verifiable rewards associated with precise logic following}. The radar chart in Fig.\ref{fig:metric} highlights the significant performance gains of Qwen3-1.7B-LogicIF over base model across both in-domain and out-of-domain benchmarks: Qwen3-1.7B-LogicIF has a substantial gain of +16.7 points (Problem-level "Both" Accuracy) on \bench{} and more impressively, the model exhibits strong generalization to out-of-domain tasks, like ZebraLogic (+31.4), MATH-500 (+4.3), GPQA-Diamond (+3.1) and HumanEval (+4.6), suggesting that \textbf{mastering precise logic instruction following provides a foundational boost to complex cognitive tasks}. Specifically, we attribute these improvements to two fundamental factors: 1) the output + state trackers rewarding mechanism creates a rigorous feedback that forces the model to abandon "logical shortcuts" and instead faithfully simulate every internal execution step and 2) the mastery of universal logic structures, such as loops, nested conditions, and recursion, provides the "atomic operations" essential for all high-order cognitive tasks, enabling the model to generalize its systematic reasoning to diverse domains. Due to computational constraints, we leave the training and analysis of larger models to future work.

% Moreover, domains that are more similar to \task{} like logic reasoning (ZebraLogic) benefit more from training on \task{}.

\section{Related Work}\label{sec:related_work}
\textbf{Instruction Following} General instruction-following evaluation work typically focuses on instructions that impose constraints on the response format or content~\citep{Zhou2023-bp,Qin2024-cs,Pyatkin2025-zq}. These research centers on enabling models to follow instructions with multiple constraints~\citep{Jaroslawicz2025-pb,Pyatkin2025-zq}. \cite{Wen2024-vq} investigates compositional constraint following, introducing logic structures such as sequential logic and branching logic. In comparison, the logic in our benchmark are significantly more complex and diverse. Verification methods generally fall into two categories: heuristic functions or LLM-as-judge. While LLM-as-judge has been shown to correlate highly with human judgments~\citep{Qin2024-cs}, there is still room for improvement~\citep{Zeng2023-gg}. In addition to general-purpose instruction-following benchmarks, other datasets target specific scenarios or constraints, such as length control~\citep{Zhang2025-bh}, long-context settings~\citep{Wu2024-le}, or agentic scenarios~\citep{Qi2025-sn,Li2025-wb}. \citet{Yang2025-fq} explore LLMs’ ability to adhere to user intent while producing functionally accurate code. In contrast, our work uses code as the source to generate instructions, requiring LLMs to generate text, rather than code, in response. Some studies have suggested that reasoning may decrease instruction-following performance~\citep{Li2025-do,Fu2025-la}. However, our findings in Section~\ref{sec:performance} indicate that reasoning can actually enhance instruction-following for logic-rich instructions. Further research like~\cite{Tam2024-va,Qin2025-du} is needed to clarify the relationship between reasoning and instruction following. 

% To the best of our knowledge, we are the first to investigate whether LLMs can precisely follow logic-rich instructions.

\textbf{Code Execution} The work on code execution evaluates LLMs by giving them code snippets and asking them to act as interpreters that predict execution outcomes~\cite{La-Malfa2024-ky,Gu2024-ro,Jain2024-xl,La-Malfa2025-nr,Sun2025-mz}. In contrast, our goal is to assess whether LLMs can follow complex logical instructions expressed in natural language. Code functions in our framework are not the input modality; they serve only as a scaffolding to synthesize natural-language instructions with rich logical structures. Models are then evaluated on whether they can faithfully execute these instructions step-by-step in natural language. This capability is of broad practical relevance, as modern agentic workflows and system constraints are primarily communicated in natural language and often encode intricate conditional reasoning, error handling, or function-calling behaviors. Our benchmark directly targets this ability.

\blue{\textbf{}}

\section{Conclusion \& Future Work}
In this paper, we propose Logic Instruction Following, a task that requires models to precisely follow and simulate each logical step within complex instructions. We further introduce a data synthesis framework, \framework{}, which automatically generates verifiable, task-intensive, and logic-rich instructions from code functions. Using \framework{}, we construct \bench{} and \trainset{} for evaluating and training LLMs on complex logic instruction following. Our experiments reveal a significant deficiency in the ability of both proprietary and open-source models to follow such logic-intensive instructions. Moreover, \trainset{} proves to be an effective resource for improving this capability, with gains that generalize to related abilities such as reasoning and coding.

% We also present \bench{}, a challenging instruction-following evaluation benchmark constructed using \framework{}, which consists of 426 tasks featuring complex logic. Experiments show that most proprietary and open-source models struggle to solve these tasks, revealing a significant deficiency in instruction-following capabilities. Future work includes exploring the use of \framework{} as a verifiable instruction generator for model training and \bench{} for evaluation to develop models with generalized and robust instruction-following capabilities. We believe that as LLMs are deployed in diverse agentic tasks, instruction following is the most crucial, yet often overlooked, capability to handle the growing complexity of memory and context managing~\citep{Gutierrez2024-ju,Khattab2023-sx}, tool calling~\citep{Wu2025-ow,Yin2025-qq}, reasoning, planning, and acting~\citep{Liu2025-vs}. The related work is included in Appendix Section~\ref{sec:related_work} due to space constraints.

% \section{Impact Statement}
% This paper presents work whose goal is to advance the field of machine learning, specifically by improving the reliability and precision of Large Language Models (LLMs) in following complex, logic-rich instructions. By providing a scalable framework for generating verifiable tasks and a rigorous benchmark for evaluation, we aim to facilitate the development of more robust agentic AI systems capable of handling intricate real-world workflows. There are many potential societal consequences of our work, none of which we feel must be specifically highlighted here.

\bibliography{IF}
\bibliographystyle{colm2026_conference}

\clearpage
\appendix

\section{Function Description Example}\label{apx:func_desc}
% \begin{figure}[htbp]
% \centering
% \begin{lstlisting}[language=Python,breaklines=true,basicstyle=\scriptsize
% \ttfamily]
% def f(lst):
%     mask = 0
%     arr = []
%     total_iterations = 0
%     unique_count = 0
%     for x in lst:
%         total_iterations += 1
%         bit = 1 << (x - 1)
%         if (mask & bit) == 0:
%             arr.append(x)
%             mask |= bit
%             unique_count += 1
%     return arr, {'total_iterations': total_iterations, 'unique_count': unique_count}
% \end{lstlisting}
% \caption{A code example. (Seed: CodeForces 1208D)}
% \label{fig:into_case}
% \end{figure}

\begin{tcolorbox}[
    colback=blue!2!white,
    colframe=blue!15!black,
    title=Function Description,
    fonttitle=\bfseries,
    rounded corners,
    drop shadow,
    breakable,
    left=3pt,
    right=3pt,
    fontupper=\scriptsize
]

\textbf{INPUTS:} 

You start with two pieces of data. First is a single whole number called $n$, which tells you how many items you should expect. Second is a list called $A$, which contains exactly $n$ numeric values. 

\textbf{LOGICS:} 

Okay, here's what you do step by step. First, take your list $A$ and sort it from largest down to smallest to make a new list, which we’ll call $B$. Now lay out four things on your workspace:  

\begin{enumerate}
    \item A running total called \texttt{val}, which you set to $0$.  
    \item An empty collection called \texttt{h} – you can think of it as a bag where you’ll always be able to pull out the smallest number.  
    \item A counter called \texttt{rm} for counting how many numbers you remove, which you set to $0$.  
    \item A tracker called \texttt{max\_sz} for the largest size \texttt{h} ever reaches, which you also set to $0$.  
\end{enumerate}

Now you're going to process each number in $B$ in order, from the first (which is the largest) down to the last (the smallest). For each item, do the following:  

\begin{itemize}
    \item Let $x$ be the current item from $B$.  
    \item Add $x$ to your running total \texttt{val} (so \texttt{val = val + x}).  
    \item Drop $x$ into your bag \texttt{h}.  
    \item Check how many items are now in \texttt{h}. If that count is bigger than \texttt{max\_sz}, then write down the new count as the new \texttt{max\_sz}.  
    \item Next, look at your running total \texttt{val}. If \texttt{val} is now below zero (that means \texttt{val < 0}), you have to remove one item from \texttt{h}. To do that, find the smallest number currently sitting in \texttt{h}, call that $y$, and take it out of \texttt{h}. Then subtract $y$ from \texttt{val} (so \texttt{val = val - y}) and add $1$ to your removal counter \texttt{rm}.
\end{itemize}

After you've done this for every single $x$ in $B$, you’re finished with the loop. No more items left to process. 

\textbf{OUTPUTS:} 

When you're all done, you end up with two things. First, the number of items left in your bag \texttt{h} – that’s the first result you’ll report. Second, you have two statistics: the total \texttt{rm}, which tells you how many times you removed the smallest item, and \texttt{max\_sz}, which tells you the largest number of items that ever piled up in \texttt{h} at once.

\end{tcolorbox}

\section{\benchmini{} Performance}
\begin{figure}[h!]
    \centering
    \includegraphics[width=0.8\linewidth]{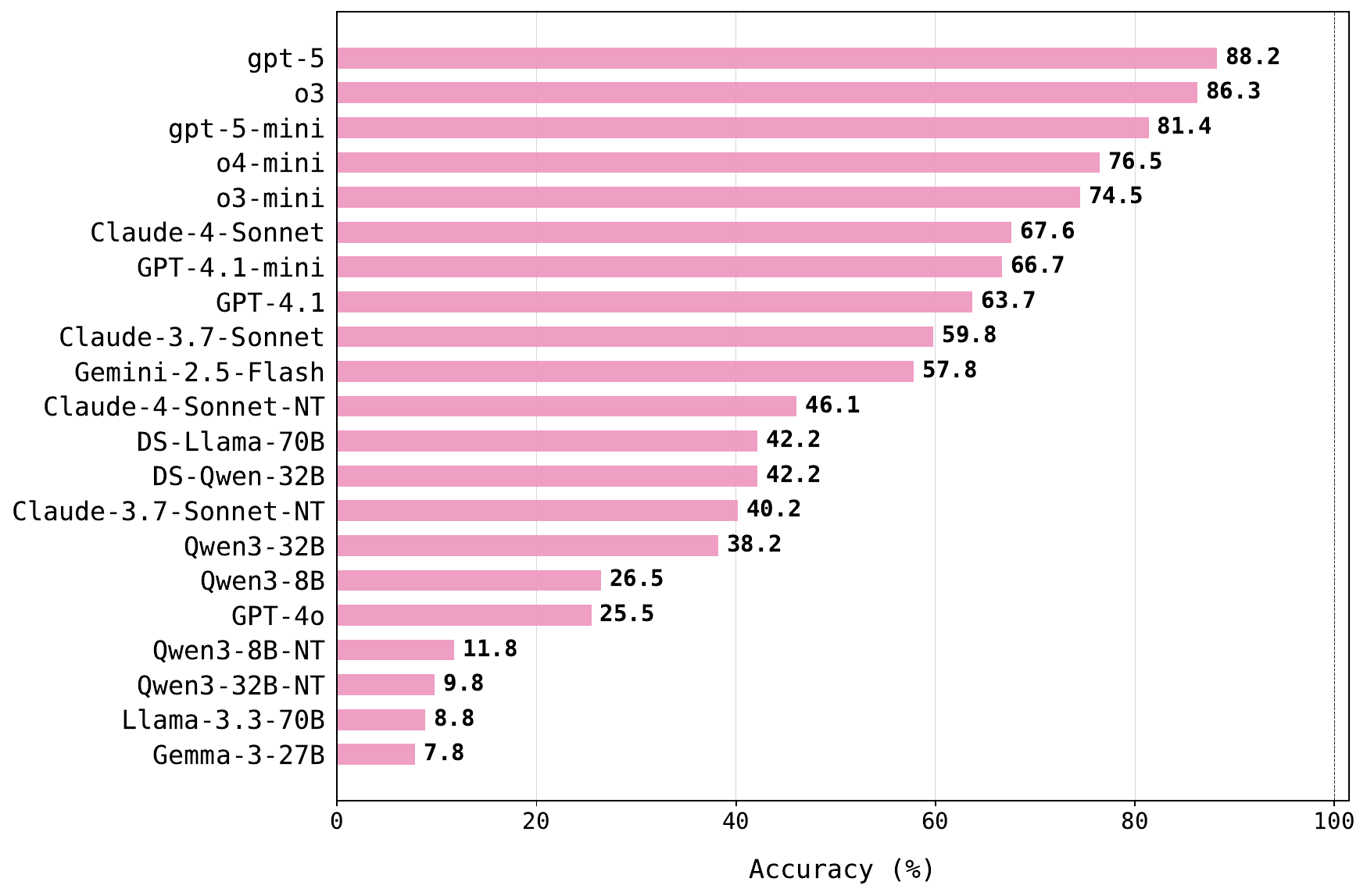}
    \caption{Overall instruction-following performance (\%) of all evaluated models on \benchmini{}, measured by the percentage of test cases where both the output and state trackers match the ground truth. Models are sorted by overall accuracy.}
    \label{fig:overall_mini}
\end{figure}

\clearpage
\section{Error Cases}\label{error_case}
% Error examples for paper
% Generated automatically from error analysis

\subsection{Error Case: POJ-1852}

\textbf{Error Type:} Control Flow Misexecution

\textbf{Test Input:}
\begin{lstlisting}[breaklines=true,basicstyle=\scriptsize\ttfamily]
4, [[2, 1, 2], [2, 2, 3], [1, 1, 3], [2, 3, 1], [1, 4, 4]]
\end{lstlisting}

\textbf{LLM Results:}
\begin{lstlisting}[breaklines=true,basicstyle=\scriptsize\ttfamily]
output: 1
stats: {'find_calls': 12, 'unions': 2}
\end{lstlisting}

\textbf{Code Results:}
\begin{lstlisting}[breaklines=true,basicstyle=\scriptsize\ttfamily]
output: 1
stats: {'find_calls': 13, 'unions': 2}
\end{lstlisting}

\textbf{Function:}
\begin{lstlisting}[language=Python,breaklines=true,basicstyle=\scriptsize\ttfamily]
def f(n, L):
    A = list(range(n+1))
    B = [0]*(n+1)
    find_calls = 0
    unions = 0
    def g(x):
        nonlocal find_calls
        find_calls += 1
        if A[x] != x:
            p = A[x]
            r = g(p)
            B[x] = (B[x] + B[p]) % 3
            A[x] = r
        return A[x]
    m = 0
    for d, x, y in L:
        if x < 1 or x > n or y < 1 or y > n:
            m += 1
            continue
        r = 0 if d == 1 else 1
        rx = g(x)
        ry = g(y)
        if rx == ry:
            if (B[x] - B[y] - r) % 3 != 0:
                m += 1
        else:
            unions += 1
            delta = (B[y] - B[x] + r) % 3
            A[rx] = ry
            B[rx] = delta
    return m, {'find_calls': find_calls, 'unions': unions}
\end{lstlisting}

\begin{tcolorbox}[
    colback=blue!5!white,
    colframe=blue!25!black,
    title=Function Description,
    fonttitle=\bfseries,
    fontupper=\scriptsize,
    rounded corners,
    drop shadow,
    breakable,
    left=3pt,
    right=3pt
]
\textbf{INPUTS:} You start with a single integer \texttt{n} and a list \texttt{L}. The list \texttt{L} contains multiple items, and each item is a triple of integers written as $(d, x, y)$. Those are all the pieces you’ll need to begin. \\

\textbf{LOGICS:} First, what you’re going to do is build two lists of length $n+1$. Call the first one \texttt{A} and fill it so that $\texttt{A}[0]=0$, $\texttt{A}[1]=1$, $\texttt{A}[2]=2$, all the way up to $\texttt{A}[n]=n$. Call the second one \texttt{B} and fill every slot from $\texttt{B}[0]$ through $\texttt{B}[n]$ with zero. Then make two counters and set both to zero: one called \texttt{find\_calls} and the other called \texttt{unions}. \\

Next you define a little helper routine, let’s say it’s called \texttt{g(x)}. Here’s exactly how you run \texttt{g} on some index $x$: 
\begin{enumerate}
  \item Right when you enter \texttt{g}, add 1 to \texttt{find\_calls}. 
  \item Check if $\texttt{A}[x]$ is the same number as $x$. If it is, you’ve found a root and you just return $x$ immediately. 
  \item If $\texttt{A}[x]$ is not $x$, that means $x$ points up to a parent. So write down $p = \texttt{A}[x]$. Then call \texttt{g(p)} to chase up the chain---which itself bumps \texttt{find\_calls} again and eventually returns a root $r$. 
  \item After \texttt{g(p)} returns $r$, you adjust $\texttt{B}[x]$: take the old $\texttt{B}[x]$, add $\texttt{B}[p]$, then reduce that sum modulo 3 (so you do $(\texttt{B}[x] + \texttt{B}[p]) \bmod 3$) and store it back into $\texttt{B}[x]$. 
  \item Also update $\texttt{A}[x]$ so that it now directly equals $r$. 
  \item Finally return $r$.
\end{enumerate}

All set up now? Good. Next you create a counter \texttt{m} and set it to zero. This will count how many triples get tossed or prove contradictory. Then you go through each triple in your list \texttt{L} in order. For each triple do this:
\begin{itemize}
  \item First look at $x$ and $y$ and see if either one is below 1 or above $n$. If $x < 1$ or $x > n$ or $y < 1$ or $y > n$, add 1 to \texttt{m} and skip the rest of these steps for that triple. (That’s just a bad-index check.)
  \item Otherwise, you figure out a small value $r$: if $d$ equals 1 then $r = 0$, else $r = 1$. 
  \item Then call \texttt{g(x)} and call \texttt{g(y)}. Store the results as \texttt{rx} and \texttt{ry}. Those are the roots of $x$ and $y$, and by now \texttt{find\_calls} has grown by two. 
  \item Now check: if \texttt{rx} and \texttt{ry} are the same number, that means $x$ and $y$ are already in the same group. In that case you compute $(\texttt{B}[x] - \texttt{B}[y] - r) \bmod 3$; if that result is not zero, it means this triple contradicts existing info, so add 1 to \texttt{m}. If it’s zero, do nothing extra---just move on to the next triple.
  \item If \texttt{rx} and \texttt{ry} are different, you need to merge the two groups. So do \texttt{unions} $+= 1$. Then compute $\delta = (\texttt{B}[y] - \texttt{B}[x] + r) \bmod 3$. After that, link the root \texttt{rx} under \texttt{ry} by setting $\texttt{A}[\texttt{rx}] = \texttt{ry}$, and store $\texttt{B}[\texttt{rx}] = \delta$. That finishes merging for this triple.
  \item Then move on to the next triple until you exhaust the list.
\end{itemize}

When you’ve processed every triple in \texttt{L}, you stop. The number \texttt{m} reflects how many triples were either invalid or contradictory, \texttt{find\_calls} tells you how many times you ran the helper \texttt{g}, and \texttt{unions} tells you how many merges you performed. \\

\textbf{OUTPUTS:} At the end you get two things. First you get the integer \texttt{m}, which is the count of all the bad or contradictory triples you hit. Second you get a small record or dictionary with two entries: \texttt{'find\_calls'} is the total number of times you entered the helper \texttt{g}, and \texttt{'unions'} is how many times you merged two distinct roots.
\end{tcolorbox}

\begin{tcolorbox}[
    colback=gray!3!white,
    colframe=gray!30!black,
    title=LLM Reasoning,
    fonttitle=\bfseries,
    fontupper=\scriptsize,
    rounded corners,
    drop shadow,
    breakable
]
\textbf{Reasoning:}

Let me work through this algorithm step by step.

\vspace{0.3em}
\textbf{Initialization:}
\begin{itemize}
    \item \texttt{n} = 4, \texttt{L} = [[2, 1, 2], [2, 2, 3], [1, 1, 3], [2, 3, 1], [1, 4, 4]]
    \item \texttt{A} = [0, 1, 2, 3, 4] (indices 0 through 4)
    \item \texttt{B} = [0, 0, 0, 0, 0] (all zeros)
    \item \texttt{find\_calls} = 0, \texttt{unions} = 0, \texttt{m} = 0
\end{itemize}

\vspace{0.3em}
\textbf{Processing Triple 1: [2, 1, 2]}
\begin{itemize}
    \item $d=2$, $x=1$, $y=2$. Both in bounds $[1,4]$.
    \item $r = 1$ (since $d \neq 1$)
    \item $g(1)$: \texttt{find\_calls} = 1, $\texttt{A}[1] = 1$ so return $1$. $r_x = 1$
    \item $g(2)$: \texttt{find\_calls} = 2, $\texttt{A}[2] = 2$ so return $2$. $r_y = 2$
    \item $r_x \neq r_y$, so merge: \texttt{unions} = 1
    \item $\texttt{delta} = (\texttt{B}[2] - \texttt{B}[1] + 1) \bmod 3 = (0 - 0 + 1) \bmod 3 = 1$
    \item $\texttt{A}[1] = 2$, $\texttt{B}[1] = 1$
\end{itemize}

\vspace{0.3em}
\textbf{Processing Triple 2: [2, 2, 3]}
\begin{itemize}
    \item $d=2$, $x=2$, $y=3$. Both in bounds.
    \item $r = 1$
    \item $g(2)$: \texttt{find\_calls} = 3, $\texttt{A}[2] = 2$ so return $2$. $r_x = 2$
    \item $g(3)$: \texttt{find\_calls} = 4, $\texttt{A}[3] = 3$ so return $3$. $r_y = 3$
    \item $r_x \neq r_y$, so merge: \texttt{unions} = 2
    \item $\texttt{delta} = (\texttt{B}[3] - \texttt{B}[2] + 1) \bmod 3 = (0 - 0 + 1) \bmod 3 = 1$
    \item $\texttt{A}[2] = 3$, $\texttt{B}[2] = 1$
\end{itemize}

\vspace{0.3em}
\textbf{Processing Triple 3: [1, 1, 3]}
\begin{itemize}
    \item $d=1$, $x=1$, $y=3$. Both in bounds.
    \item $r = 0$
    \item $g(1)$: \texttt{find\_calls} = 5, $\texttt{A}[1] = 2 \neq 1$, so $p=2$
    \begin{itemize}
        \item $g(2)$: \texttt{find\_calls} = 6, $\texttt{A}[2] = 3 \neq 2$, so $p=3$
        \begin{itemize}
            \item $g(3)$: \texttt{find\_calls} = 7, $\texttt{A}[3] = 3$ so return $3$
        \end{itemize}
        \item $\texttt{B}[2] = (1 + 0) \bmod 3 = 1$, $\texttt{A}[2] = 3$, return 3
        \item $\texttt{B}[1] = (1 + 1) \bmod 3 = 2$, $\texttt{A}[1] = 3$, return 3
    \end{itemize}
    \item $r_x = 3$
    \item $g(3)$: \texttt{find\_calls} = 8, $\texttt{A}[3] = 3$ so return $3$. $r_y = 3$
    \item $r_x = r_y = 3$, check consistency: $(\texttt{B}[1] - \texttt{B}[3] - 0) \bmod 3 = (2 - 0 - 0) \bmod 3 = 2 \neq 0$
    \item Contradiction! $\texttt{m} = 1$
\end{itemize}

\vspace{0.3em}
\textbf{Processing Triple 4: [2, 3, 1]}
\begin{itemize}
    \item $d=2$, $x=3$, $y=1$. Both in bounds.
    \item $r = 1$
    \item $g(3)$: \texttt{find\_calls} = 9, $\texttt{A}[3] = 3$ so return $3$. $r_x = 3$
    \item $g(1)$: \texttt{find\_calls} = 10, $\texttt{A}[1] = 3$ so return $3$. $r_y = 3$ (\red{Control Flow Misexecution (code line 9): The model should enter the if block because A[x]!=x.})
    \item $r_x = r_y = 3$, check consistency: $(\texttt{B}[3] - \texttt{B}[1] - 1) \bmod 3 = (0 - 2 - 1) \bmod 3 = (-3) \bmod 3 = 0$
    \item Consistent, no increment to \texttt{m}
\end{itemize}

\vspace{0.3em}
\textbf{Processing Triple 5: [1, 4, 4]}
\begin{itemize}
    \item $d=1$, $x=4$, $y=4$. Both in bounds.
    \item $r = 0$
    \item $g(4)$: \texttt{find\_calls} = 11, $\texttt{A}[4] = 4$ so return $4$. $r_x = 4$
    \item $g(4)$: \texttt{find\_calls} = 12, $\texttt{A}[4] = 4$ so return $4$. $r_y = 4$
    \item $r_x = r_y = 4$, check consistency: $(\texttt{B}[4] - \texttt{B}[4] - 0) \bmod 3 = (0 - 0 - 0) \bmod 3 = 0$
    \item Consistent, no increment to \texttt{m}
\end{itemize}

\vspace{0.5em}
\textbf{Output:} $1$

\vspace{0.3em}
\textbf{Statistics:} \texttt{\{'find\_calls': 12, 'unions': 2\}}
\end{tcolorbox}

\clearpage
% Error examples for paper
% Generated automatically from error analysis

\subsection{Error Case: CodeForces-328A}

\textbf{Error Type:} Instruction Misinterpretation

\textbf{Test Input:}
\begin{lstlisting}[breaklines=true,basicstyle=\scriptsize\ttfamily]
4, [1, 0, 1, 0]
\end{lstlisting}

\textbf{LLM Results:}
\begin{lstlisting}[breaklines=true,basicstyle=\scriptsize\ttfamily]
output: 3
stats: {'rec_calls': 7, 'max_depth': 5}
\end{lstlisting}

\textbf{Code Results:}
\begin{lstlisting}[breaklines=true,basicstyle=\scriptsize\ttfamily]
output: 3
stats: {'rec_calls': 7, 'max_depth': 3}
\end{lstlisting}

\textbf{Function:}
\begin{lstlisting}[language=Python,breaklines=true,basicstyle=\scriptsize\ttfamily]
def f(n, arr):
    c1 = 0
    c2 = 0
    x = 0
    for val in arr:
        x += val
    A = [1 if val == 0 else -1 for val in arr]
    def g(l, r, depth):
        nonlocal c1, c2
        c1 += 1
        if depth > c2:
            c2 = depth
        if l == r:
            v = A[l]
            return v, v, v, v
        m = (l + r) // 2
        lt, lp, ls, lb = g(l, m, depth + 1)
        rt, rp, rs, rb = g(m + 1, r, depth + 1)
        total = lt + rt
        pref = max(lp, lt + rp)
        suf = max(rs, rt + ls)
        best = max(lb, rb, ls + rp)
        return total, pref, suf, best
    _, _, _, b = g(0, n - 1, 1)
    return x + b, {'rec_calls': c1, 'max_depth': c2}
\end{lstlisting}

\begin{tcolorbox}[
    colback=blue!5!white,
    colframe=blue!25!black,
    title=Function Description,
    fonttitle=\bfseries,
    fontupper=\scriptsize,
    rounded corners,
    drop shadow,
    breakable,
    left=3pt,
    right=3pt
]
\textbf{INPUTS:} The inputs are a number called \texttt{n} and a list called \texttt{arr} containing \texttt{n} numbers. \\

\textbf{LOGICS:} Start by writing down three counters named \texttt{c1}, \texttt{c2}, and \texttt{x} and set them all to zero. Now go through each value in \texttt{arr} one at a time and add that value to \texttt{x}. When you finish, \texttt{x} holds the sum of \texttt{arr}. \\

Next, build a new list called \texttt{A}: begin with it empty, then for every element in \texttt{arr} in order, if the element is zero append 1 into \texttt{A}, otherwise append $-1$. \\

Now you are going to define a process named \texttt{g} that takes three numbers \texttt{l}, \texttt{r}, and \texttt{depth}. Every time you start \texttt{g}, do the following: first add 1 to \texttt{c1}, then compare \texttt{depth} with \texttt{c2} and if \texttt{depth} is larger replace \texttt{c2} with \texttt{depth}. Then check if \texttt{l} equals \texttt{r}. If they are equal, look up \texttt{A} at index \texttt{l}, call that value \texttt{v}, and return four copies of \texttt{v}. \\

If \texttt{l} is less than \texttt{r}, compute \texttt{m} as the floor of $(\texttt{l} + \texttt{r})$ divided by 2. Now call \texttt{g} on the left half by using \texttt{l}, \texttt{m}, and \texttt{depth} plus 1, and write down its four outputs as \texttt{lt}, \texttt{lp}, \texttt{ls}, and \texttt{lb} in that order. Then call \texttt{g} on the right half by using \texttt{m} + 1, \texttt{r}, and \texttt{depth} plus 1, and write down its four outputs as \texttt{rt}, \texttt{rp}, \texttt{rs}, and \texttt{rb}. \\

After you have those eight numbers, calculate \texttt{total} as \texttt{lt} plus \texttt{rt}, then \texttt{pref} as whichever is larger between \texttt{lp} and \texttt{lt} plus \texttt{rp}, then \texttt{suf} as whichever is larger between \texttt{rs} and \texttt{rt} plus \texttt{ls}, and finally \texttt{best} as the largest of \texttt{lb}, \texttt{rb}, and \texttt{ls} plus \texttt{rp}. Then return \texttt{total}, \texttt{pref}, \texttt{suf}, and \texttt{best} from this call. \\

Once you have defined that, kick off the process by calling \texttt{g} with \texttt{l} = 0, \texttt{r} = \texttt{n} minus 1, and \texttt{depth} = 1. When it finishes, ignore the first three values it gives you and call the fourth value \texttt{b}. At this point \texttt{c1} is the total count of \texttt{g} calls, and \texttt{c2} is the deepest level you reached. Finally, add \texttt{b} to \texttt{x} and that gives you the main result. \\

\textbf{OUTPUTS:} You will end up with two things: a number equal to \texttt{x} plus \texttt{b}, and a small object with two fields, \texttt{rec\_calls} set to \texttt{c1} (the total number of \texttt{g} calls) and \texttt{max\_depth} set to \texttt{c2} (the deepest recursion level you reached).
\end{tcolorbox}

\begin{tcolorbox}[
    colback=gray!3!white,
    colframe=gray!30!black,
    title=LLM Reasoning,
    fonttitle=\bfseries,
    fontupper=\scriptsize,
    rounded corners,
    drop shadow,
    breakable
]
\textbf{Reasoning:}

Let me work through this algorithm step by step.

\medskip
\textbf{Step 1: Initialize variables}
\begin{itemize}
    \item $n = 4$, $\texttt{arr} = [1, 0, 1, 0]$
    \item $c_1 = 0$, $c_2 = 0$, $x = 0$
\end{itemize}

\medskip
\textbf{Step 2: Calculate sum of \texttt{arr}}\\
Going through each value in \texttt{arr} and adding to $x$:
\begin{itemize}
    \item $x = 0 + 1 = 1$
    \item $x = 1 + 0 = 1$
    \item $x = 1 + 1 = 2$\\
    \item $x = 2 + 0 = 2$
\end{itemize}
So $x = 2$

\medskip
\textbf{Step 3: Build array $A$}\\
For each element in \texttt{arr}: if 0 append 1, otherwise append $-1$:
\begin{itemize}
    \item $\texttt{arr}[0] = 1 \rightarrow$ append $-1$
    \item $\texttt{arr}[1] = 0 \rightarrow$ append $1$
    \item $\texttt{arr}[2] = 1 \rightarrow$ append $-1$\\
    \item $\texttt{arr}[3] = 0 \rightarrow$ append $1$
\end{itemize}
So $A = [-1, 1, -1, 1]$

\medskip
\textbf{Step 4: Execute $g(0, 3, 1)$}

\medskip
\textbf{$g(0, 3, 1)$:}
\begin{itemize}
    \item $c_1 = 1$, depth$=1 > c_2=0$ so $c_2 = 1$
    \item $l \neq r$, so $m = \lfloor (0+3)/2 \rfloor = 1$
    \item Call $g(0, 1, 2)$ and $g(2, 3, 3)$ (\red{Instruction Misinterpretation (code line 18): The second function call should be $g(2, 3, 2)$.})
\end{itemize}

\medskip
\textbf{$g(0, 1, 2)$:}
\begin{itemize}
    \item $c_1 = 2$, depth$=2 > c_2=1$ so $c_2 = 2$
    \item $l \neq r$, so $m = \lfloor (0+1)/2 \rfloor = 0$
    \item Call $g(0, 0, 3)$ and $g(1, 1, 4)$
\end{itemize}

\medskip
\textbf{$g(0, 0, 3)$:}
\begin{itemize}
    \item $c_1 = 3$, depth$=3 > c_2=2$ so $c_2 = 3$
    \item $l = r = 0$, so $v = A[0] = -1$
    \item Return $(-1, -1, -1, -1)$
\end{itemize}

\medskip
\textbf{$g(1, 1, 4)$:}
\begin{itemize}
    \item $c_1 = 4$, depth$=4 > c_2=3$ so $c_2 = 4$
    \item $l = r = 1$, so $v = A[1] = 1$
    \item Return $(1, 1, 1, 1)$
\end{itemize}

Back to $g(0, 1, 2)$:
\[
\begin{cases}
    lt = -1, \quad lp = -1, \quad ls = -1, \quad lb = -1; \\
    rt = 1, \quad rp = 1, \quad rs = 1, \quad rb = 1
\end{cases}
\]
\[
\begin{aligned}
    \text{total} &= -1 + 1 = 0 \\
    \text{pref} &= \max(-1, -1 + 1) = 0 \\
    \text{suf} &= \max(1, 1 + (-1)) = 1 \\
    \text{best} &= \max(-1, 1, -1 + 1) = 1
\end{aligned}
\]
Return $(0, 0, 1, 1)$

\medskip
\textbf{$g(2, 3, 3)$:}
\begin{itemize}
    \item $c_1 = 5$, depth$=3 \leq c_2=4$ so $c_2$ stays 4
    \item $l \neq r$, so $m = \lfloor (2+3)/2 \rfloor = 2$
    \item Call $g(2, 2, 4)$ and $g(3, 3, 5)$
\end{itemize}

\medskip
\textbf{$g(2, 2, 4)$:}
\begin{itemize}
    \item $c_1 = 6$, depth$=4 = c_2=4$ so $c_2$ stays 4
    \item $l = r = 2$, so $v = A[2] = -1$
    \item Return $(-1, -1, -1, -1)$
\end{itemize}

\medskip
\textbf{$g(3, 3, 5)$:}
\begin{itemize}
    \item $c_1 = 7$, depth$=5 > c_2=4$ so $c_2 = 5$
    \item $l = r = 3$, so $v = A[3] = 1$
    \item Return $(1, 1, 1, 1)$
\end{itemize}

Back to $g(2, 3, 3)$:
\[
\begin{cases}
    lt = -1, \quad lp = -1, \quad ls = -1, \quad lb = -1; \\
    rt = 1, \quad rp = 1, \quad rs = 1, \quad rb = 1
\end{cases}
\]
\[
\begin{aligned}
    \text{total} &= -1 + 1 = 0 \\
    \text{pref} &= \max(-1, -1 + 1) = 0 \\
    \text{suf} &= \max(1, 1 + (-1)) = 1 \\
    \text{best} &= \max(-1, 1, -1 + 1) = 1
\end{aligned}
\]
Return $(0, 0, 1, 1)$

\medskip
Back to $g(0, 3, 1)$:
\[
\begin{cases}
    lt = 0, \quad lp = 0, \quad ls = 1, \quad lb = 1; \\
    rt = 0, \quad rp = 0, \quad rs = 1, \quad rb = 1
\end{cases}
\]
\[
\begin{aligned}
    \text{total} &= 0 + 0 = 0 \\
    \text{pref} &= \max(0, 0 + 0) = 0 \\
    \text{suf} &= \max(1, 0 + 1) = 1 \\
    \text{best} &= \max(1, 1, 1 + 0) = 1
\end{aligned}
\]
Return $(0, 0, 1, 1)$

\medskip
\textbf{Step 5: Final calculations}
\begin{itemize}
    \item $b = 1$ (fourth value from $g$)
    \item $c_1 = 7$ (total $g$ calls)
    \item $c_2 = 5$ (max depth)
    \item Final result $= x + b = 2 + 1 = 3$
\end{itemize}

\medskip
\textbf{Output:} 3

\medskip
\textbf{Statistics:} \{\texttt{'rec\_calls'}: 7, \texttt{'max\_depth'}: 5\}
\end{tcolorbox}

\clearpage

% Error examples for paper
% Generated automatically from error analysis

\subsection{Error Case: CODEFORCES-2018B}

\textbf{Error Type:} Misordered Execution

\textbf{Test Input:}
\begin{lstlisting}[breaklines=true,basicstyle=\scriptsize\ttfamily]
8, [13, 27, 35, ...] (len=8)
\end{lstlisting}

\textbf{LLM Results:}
\begin{lstlisting}[breaklines=true,basicstyle=\scriptsize\ttfamily]
output: 21
stats: {'xor_operations': 7, 'and_operations': 7}
\end{lstlisting}

\textbf{Code Results:}
\begin{lstlisting}[breaklines=true,basicstyle=\scriptsize\ttfamily]
output: 24
stats: {'xor_operations': 8, 'and_operations': 7}
\end{lstlisting}

\textbf{Function:}
\begin{lstlisting}[language=Python,breaklines=true,basicstyle=\scriptsize\ttfamily]
def f(n, arr):
    m = n - 1
    val = 0
    x = m
    y = 0
    z = 0
    while True:
        val ^= arr[x]
        y += 1
        if x == 0:
            break
        x = (x - 1) & m
        z += 1
    return val, {'xor_operations': y, 'and_operations': z}
\end{lstlisting}

\begin{tcolorbox}[
    colback=blue!5!white,
    colframe=blue!25!black,
    title=Function Description,
    fonttitle=\bfseries,
    fontupper=\scriptsize,
    rounded corners,
    drop shadow,
    breakable,
    left=3pt,
    right=3pt
]
\textbf{INPUTS:} You're working with two pieces of data: a number called \texttt{n}, and a list called \texttt{arr} that contains exactly \texttt{n} numeric entries, indexed from 0 up to \texttt{n} minus one.\\[6pt]

\textbf{LOGICS:} Start by creating a variable called \texttt{m} by taking \texttt{n} and subtracting 1. Next, create a running total named \texttt{val} and set it to zero. Now set up another variable \texttt{x} and give it the same value as \texttt{m}. Also prepare two counters, \texttt{y} and \texttt{z}, and initialize both of them at zero. What you're going to do now is enter a loop that keeps going until \texttt{x} becomes zero. Every time you go through this loop, follow these exact steps: \\
\begin{enumerate}
    \item First, look up the item at index \texttt{x} in your \texttt{arr} list and apply a bitwise exclusive OR between that item and your current \texttt{val}. To do that by hand, write both numbers in binary, compare each bit pair, write down a 1 if the bits differ or a 0 if they match, then turn that resulting bit pattern back into a number and call it your new \texttt{val}.
    \item Once you've done that, add 1 to your counter \texttt{y} to record that XOR operation.
    \item Next, check whether \texttt{x} is equal to zero. If \texttt{x} is zero, break out of the loop immediately---you're done.
    \item If \texttt{x} is not zero, you need to update \texttt{x} for the next round: subtract 1 from your current \texttt{x}, then take that result and apply a bitwise AND with \texttt{m} (again, line up the binary bits of \texttt{(x-1)} and \texttt{m}, write down a 1 only where both bits are 1, then convert that back into a number). Store that outcome back into \texttt{x}, and because you just performed an AND, add 1 to your counter \texttt{z}.
    \item Then go back to the top of the loop and repeat exactly those steps---XOR and increment \texttt{y}, check for \texttt{x == 0} and maybe break, AND and increment \texttt{z}---until \texttt{x} finally reaches zero and you exit.
\end{enumerate}

\textbf{OUTPUTS:} When you're finished, you'll have two things: the first is the final number stored in \texttt{val} after all the XOR steps, and the second is a small dictionary showing \texttt{\{'xor\_operations': y, 'and\_operations': z\}} which equals your \texttt{y} counter (the total number of XORs you performed), and \texttt{z} counter (the total number of ANDs you performed).
\end{tcolorbox}

\begin{tcolorbox}[
    colback=gray!3!white,
    colframe=gray!30!black,
    title=LLM Reasoning,
    fonttitle=\bfseries,
    fontupper=\scriptsize,
    rounded corners,
    drop shadow,
    breakable
]
\textbf{Reasoning:}

Given the input value: $n = 8$, and \texttt{arr} = [13, 27, 35, 49, 59, 61, 73, 83].

\vspace{0.5em}
\textbf{Step 1: Initialization}
\begin{itemize}
    \item Compute $m = n - 1 = 8 - 1 = 7$ (binary: 0111)
    \item Initialize \texttt{val} = 0
    \item $x = m = 7$
    \item $y$ (XOR counter) = 0
    \item $z$ (AND counter) = 0
\end{itemize}

\vspace{0.5em}
\textbf{Step 2: Loop until $x == 0$}

\medskip
\hrule
\medskip

\textbf{Iteration 1: $x = 7$}
\begin{itemize}
    \item \texttt{arr[7]} = 83  
    \item XOR: \texttt{val} = 0 XOR 83  
    \begin{itemize}
        \item Convert to binary: 0 (00000000) XOR 83 (01010011) = 83 (01010011)  
    \end{itemize}
    \item $y \mathrel{+}= 1 \to y = 1$  
    \item $x$ is not 0 $\to$ proceed  
    \item $x = (x - 1) \land m = (7 - 1 = 6) \land 7 = 6$ (binary: 0110 AND 0111 = 0110)  
    \item $z \mathrel{+}= 1 \to z = 1$  
\end{itemize}

\textbf{Iteration 2: $x = 6$}
\begin{itemize}
    \item \texttt{arr[6]} = 73  
    \item XOR: \texttt{val} = 83 XOR 73  
    \begin{itemize}
        \item 83: 01010011  
        \item 73: 01001001  
        \item XOR: 00011010 $\to$ decimal: 26  
    \end{itemize}
    \item $y \mathrel{+}= 1 \to y = 2$  
    \item $x$ is not 0 $\to$ proceed  
    \item $x = (6 - 1) \land 7 = 5 \land 7 = 5$ (binary: 0101 AND 0111 = 0101)  
    \item $z \mathrel{+}= 1 \to z = 2$  
\end{itemize}

\textbf{Iteration 3: $x = 5$}
\begin{itemize}
    \item \texttt{arr[5]} = 61  
    \item XOR: \texttt{val} = 26 XOR 61  
    \begin{itemize}
        \item 26: 00011010  
        \item 61: 00111101  
        \item XOR: 00100111 $\to$ decimal: 39  
    \end{itemize}
    \item $y \mathrel{+}= 1 \to y = 3$  
    \item $x$ is not 0 $\to$ proceed  
    \item $x = (5 - 1) \land 7 = 4 \land 7 = 4$ (binary: 0100 AND 0111 = 0100)  
    \item $z \mathrel{+}= 1 \to z = 3$  
\end{itemize}

\textbf{Iteration 4: $x = 4$}
\begin{itemize}
    \item \texttt{arr[4]} = 59  
    \item XOR: \texttt{val} = 39 XOR 59  
    \begin{itemize}
        \item 39: 00100111  
        \item 59: 00111011  
        \item XOR: 00011100 $\to$ decimal: 28  
    \end{itemize}
    \item $y \mathrel{+}= 1 \to y = 4$  
    \item $x$ is not 0 $\to$ proceed  
    \item $x = (4 - 1) \land 7 = 3 \land 7 = 3$ (binary: 0011 AND 0111 = 0011)  
    \item $z \mathrel{+}= 1 \to z = 4$  
\end{itemize}

\textbf{Iteration 5: $x = 3$}
\begin{itemize}
    \item \texttt{arr[3]} = 49  
    \item XOR: \texttt{val} = 28 XOR 49  
    \begin{itemize}
        \item 28: 00011100  
        \item 49: 00110001  
        \item XOR: 00101101 $\to$ decimal: 45  
    \end{itemize}
    \item $y \mathrel{+}= 1 \to y = 5$  
    \item $x$ is not 0 $\to$ proceed  
    \item $x = (3 - 1) \land 7 = 2 \land 7 = 2$ (binary: 0010 AND 0111 = 0010)  
    \item $z \mathrel{+}= 1 \to z = 5$  
\end{itemize}

\textbf{Iteration 6: $x = 2$}
\begin{itemize}
    \item \texttt{arr[2]} = 35  
    \item XOR: \texttt{val} = 45 XOR 35  
    \begin{itemize}
        \item 45: 00101101  
        \item 35: 00100011  
        \item XOR: 00001110 $\to$ decimal: 14  
    \end{itemize}
    \item $y \mathrel{+}= 1 \to y = 6$  
    \item $x$ is not 0 $\to$ proceed  
    \item $x = (2 - 1) \land 7 = 1 \land 7 = 1$ (binary: 0001 AND 0111 = 0001)  
    \item $z \mathrel{+}= 1 \to z = 6$  
\end{itemize}

\textbf{Iteration 7: $x = 1$}
\begin{itemize}
    \item \texttt{arr[1]} = 27  
    \item XOR: \texttt{val} = 14 XOR 27  
    \begin{itemize}
        \item 14: 00001110  
        \item 27: 00011011  
        \item XOR: 00010101 $\to$ decimal: 21  
    \end{itemize}
    \item $y \mathrel{+}= 1 \to y = 7$  
    \item $x$ is not 0 $\to$ proceed  
    \item $x = (1 - 1) \land 7 = 0 \land 7 = 0$ (binary: 0000 AND 0111 = 0000)  
    \item $z \mathrel{+}= 1 \to z = 7$  
\end{itemize}

\textbf{Iteration 8: $x = 0$}  
\begin{itemize}
    \item Check if $x == 0$ $\to$ yes, exit loop (\red{Misordered Execution (code line 8 9 10): The model forgets updating \texttt{val} (line 8) and increasing \texttt{y} (line 9).})
\end{itemize}

\medskip
\hrule
\medskip

\textbf{Step 3: Compute final results:}  
\begin{itemize}
    \item Final \texttt{val} = 21  
    \item \texttt{xor\_operations} ($y$) = 7  
    \item \texttt{and\_operations} ($z$) = 7  
\end{itemize}

\medskip
\hrule
\medskip

\textbf{Output:}
\begin{lstlisting}[language=python]
{
  'val': 21,
  'and_operations': 7,
  'xor_operations': 7
}
\end{lstlisting}

\textbf{Statistics:}
\begin{lstlisting}[language=python]
{'xor_operations': 7, 'and_operations': 7}
\end{lstlisting}
\end{tcolorbox}
\clearpage

% Error examples for paper
% Generated automatically from error analysis

\subsection{Error Case: CODEFORCES-1146E}

\textbf{Error Type:} Missing Logic elements

\textbf{Test Input:}
\begin{lstlisting}[breaklines=true,basicstyle=\scriptsize\ttfamily]
5, [1, 2, 3, 4, 5], [[1, 5], [2, 4], [3, 3]]
\end{lstlisting}

\textbf{LLM Results:}
\begin{lstlisting}[breaklines=true,basicstyle=\scriptsize\ttfamily]
output: [0, 1, 1]
stats: {'stack_pops': 4, 'bit_ops': 9}
\end{lstlisting}

\textbf{Code Results:}
\begin{lstlisting}[breaklines=true,basicstyle=\scriptsize\ttfamily]
output: [5, 3, 1]
stats: {'stack_pops': 4, 'bit_ops': 11}
\end{lstlisting}

\textbf{Function:}
\begin{lstlisting}[language=Python,breaklines=true,basicstyle=\scriptsize\ttfamily]
def f(n, arr, lst):
    pop_cnt = 0
    bit_ops = 0
    A = [0] * n
    L = []
    for i in range(n):
        while L and arr[L[-1]] < arr[i]:
            L.pop()
            pop_cnt += 1
        A[i] = L[-1] + 1 if L else 0
        L.append(i)
    A2 = sorted((A[i], i + 1) for i in range(n))
    B = sorted((x, y, idx) for idx, (x, y) in enumerate(lst))
    C = [0] * (n + 1)
    def upd(i):
        nonlocal bit_ops
        bit_ops += 1
        while i <= n:
            C[i] += 1
            i += i & -i
    def qry(i):
        nonlocal bit_ops
        bit_ops += 1
        s = 0
        while i > 0:
            s += C[i]
            i -= i & -i
        return s
    D = [0] * len(lst)
    j = 0
    for x, y, idx in B:
        while j < n and A2[j][0] < x:
            _, pos = A2[j]
            upd(pos)
            j += 1
        D[idx] = qry(y) - qry(x - 1)
    return D, {"stack_pops": pop_cnt, "bit_ops": bit_ops}
\end{lstlisting}

\begin{tcolorbox}[
    colback=blue!5!white,
    colframe=blue!25!black,
    title=Function Description,
    fonttitle=\bfseries,
    fontupper=\scriptsize,
    rounded corners,
    drop shadow,
    breakable,
    left=3pt,
    right=3pt
]
\textbf{INPUTS}: \texttt{n} is an integer, \texttt{arr} is a list of \texttt{n} numbers, \texttt{lst} is a list of pairs $(x, y)$ \\

\textbf{LOGICS}: \\
Start by preparing your workspace: set \texttt{pop\_cnt} to 0 and \texttt{bit\_ops} to 0. Then make a list \texttt{A} with \texttt{n} zeros and an empty list \texttt{L} that we'll use like a stack. \\

Now go index by index through \texttt{arr}: for $i$ from 0 up to $n - 1$, do this: \\

\hspace*{1em} while \texttt{L} is not empty and $\texttt{arr}[\texttt{L}[-1]] < \texttt{arr}[i]$, remove that last index from \texttt{L} and add 1 to \texttt{pop\_cnt}. \\

Once that popping stage is done, if \texttt{L} still has something, look at that last index (call it \texttt{top}) and set $A[i]$ to $\texttt{top} + 1$; if \texttt{L} is empty, set $A[i]$ to 0. Then append $i$ to \texttt{L}. \\

That completes filling \texttt{A}. Next what you do is build a list \texttt{A2} of pairs $(A[i], i+1)$ for each $i$ from $0$ to $n - 1$, and then sort \texttt{A2} in ascending order by the first number in each pair (and by the second if there's a tie). \\

Then take your input \texttt{lst} and turn it into a list \texttt{B} of triples: for each pair $(x, y)$ in \texttt{lst}, remember its original position \texttt{idx}, and form a triple $(x, y, \texttt{idx})$; once you have all of those, sort \texttt{B} in ascending order by $x$, then by $y$, then by \texttt{idx}. \\

Now set up an array \texttt{C} of length $n+1$ filled with zeros. We'll use two helper actions: \texttt{upd(i)} and \texttt{qry(i)}. \\

When you call \texttt{upd(i)}, add 1 to \texttt{bit\_ops}, then as long as $i \leq n$, add 1 to \texttt{C[i]} and then increase $i$ by the value of its lowest set bit (compute $i \& -i$ and add it to $i$). \\

When you call \texttt{qry(i)}, add 1 to \texttt{bit\_ops}, start a local sum $s$ at 0, and while $i > 0$, add \texttt{C[i]} to $s$ and then decrease $i$ by its lowest set bit ($i \& -i$); when that finishes return $s$. \\

After that, create a list \texttt{D} of the same length as \texttt{lst} and fill it with zeros, and also set a pointer \texttt{j} to 0. \\

Finally, go through each triple $(x, y, \texttt{idx})$ in \texttt{B} in order: \\

\hspace*{1em} while $\texttt{j} < n$ and $\texttt{A2[j][0]} < x$, take the second value from $\texttt{A2[j]}$ (call it \texttt{pos}), call \texttt{upd(pos)}, and add 1 to \texttt{j}; \\

\hspace*{1em} once that loop ends, call \texttt{qry(y)} to get the total up to $y$, call \texttt{qry(x - 1)} to get the total up to $x-1$, subtract the two results, and store that number into \texttt{D} at position \texttt{idx}. \\

Once you've done that for every triple in \texttt{B}, you're done with the processing. \\

\textbf{OUTPUTS}: \\
You end up with \texttt{D}, a list of numbers the same length as \texttt{lst}, where each entry is the count computed for the corresponding pair in \texttt{lst}. You also have a dictionary of statistics: \texttt{stack\_pops} is the total number of times you removed an index from \texttt{L} during that first scanning phase, and \texttt{bit\_ops} is the total number of times you entered either the \texttt{upd} or \texttt{qry} action.
\end{tcolorbox}

\begin{tcolorbox}[
    colback=gray!3!white,
    colframe=gray!30!black,
    title=LLM Reasoning,
    fonttitle=\bfseries,
    fontupper=\scriptsize,
    rounded corners,
    drop shadow,
    breakable
]
\textbf{Reasoning:}

We are given a function to execute manually with the input values \texttt{n = 5}, \texttt{arr = [1, 2, 3, 4, 5]}, and \texttt{lst = [(1, 5), (2, 4), (3, 3)]}.

Here’s my step-by-step execution:

---

\subsubsection*{Step 1: Initialize variables}
\begin{itemize}
    \item \texttt{pop\_cnt = 0}
    \item \texttt{bit\_ops = 0}
    \item \texttt{A = [0, 0, 0, 0, 0]} (length $n$)
    \item \texttt{L = []} (empty stack)
    \item \texttt{j} will be declared later
\end{itemize}

---

\subsubsection*{Step 2: Scan the array \texttt{arr} to fill list \texttt{A}}

We process \texttt{arr[i]} for $i = 0$ to $4$.

Let’s track changes in \texttt{L} and compute values for \texttt{A}:

\paragraph*{ $i=0$:}
\begin{itemize}
    \item \texttt{arr[0] = 1}, \texttt{L} is empty $\rightarrow$ no popping
    \item \texttt{A[0] = 0}
    \item Append 0 to \texttt{L}: \texttt{L = [0]}
\end{itemize}

\paragraph*{ $i=1$:}
\begin{itemize}
    \item $\texttt{arr[1]} = 2$, \texttt{L[-1]} = \texttt{arr[0]} = 1 $\rightarrow$ $\texttt{arr[1]} > \texttt{arr[0]}$, so pop 0 from \texttt{L}, \texttt{L = []}
    \item \texttt{pop\_cnt} $+= 1 \rightarrow \texttt{pop\_cnt} = 1$
    \item \texttt{L} is empty $\rightarrow$ \texttt{A[1] = 0}
    \item Append 1 to \texttt{L}: \texttt{L = [1]}
\end{itemize}

\paragraph*{ $i=2$:}
\begin{itemize}
    \item \texttt{arr[2] = 3}, \texttt{L[-1]} = \texttt{arr[1]} = 2 $\rightarrow$ $\texttt{arr[2]} > \texttt{arr[1]}$, so pop 1 from \texttt{L}, \texttt{L = []}
    \item \texttt{pop\_cnt} $+= 1 \rightarrow \texttt{pop\_cnt} = 2$
    \item \texttt{L} is empty $\rightarrow$ \texttt{A[2] = 0}
    \item Append 2 to \texttt{L}: \texttt{L = [2]}
\end{itemize}

\paragraph*{ $i=3$:}
\begin{itemize}
    \item \texttt{arr[3] = 4}, \texttt{L[-1]} = \texttt{arr[2]} = 3 $\rightarrow$ $\texttt{arr[3]} > \texttt{arr[2]}$, so pop 2 from \texttt{L}, \texttt{L = []}
    \item \texttt{pop\_cnt} $+= 1 \rightarrow \texttt{pop\_cnt} = 3$
    \item \texttt{L} is empty $\rightarrow$ \texttt{A[3] = 0}
    \item Append 3 to \texttt{L}: \texttt{L = [3]}
\end{itemize}

\paragraph*{ $i=4$:}
\begin{itemize}
    \item \texttt{arr[4] = 5}, \texttt{L[-1]} = \texttt{arr[3]} = 4 $\rightarrow$ $\texttt{arr[4]} > \texttt{arr[3]}$, so pop 3 from \texttt{L}, \texttt{L = []}
    \item \texttt{pop\_cnt} $+= 1 \rightarrow \texttt{pop\_cnt} = 4$
    \item \texttt{L} is empty $\rightarrow$ \texttt{A[4] = 0}
    \item Append 4 to \texttt{L}: \texttt{L = [4]}
\end{itemize}

\noindent
$\checkmark$ Now $\texttt{A} = [0, 0, 0, 0, 0]$, and $\texttt{L} = [4]$, and $\texttt{pop\_cnt} = 4$

---

\subsubsection*{Step 3: Build list \texttt{A2}}

\texttt{A2} is list of pairs: $(A[i], i+1)$, one for each $i$:

\begin{itemize}
    \item \texttt{A0} = 0 $\rightarrow (0, 1)$
    \item \texttt{A1} = 0 $\rightarrow (0, 2)$
    \item \texttt{A2} = 0 $\rightarrow (0, 3)$
    \item \texttt{A3} = 0 $\rightarrow (0, 4)$
    \item \texttt{A4} = 0 $\rightarrow (0, 5)$
\end{itemize}

Sort \texttt{A2} by first ($A[i]$) and tie-breaker ($i+1$). Since all $A[i] = 0$, sort by $i+1$:

\[
\texttt{A2 sorted} = [(0, 1), (0, 2), (0, 3), (0, 4), (0, 5)]
\]

---

\subsubsection*{Step 4: Build list \texttt{B}}

Original $\texttt{lst} = [[1, 5], [2, 4], [3, 3]]$

Assign index to each element:

\begin{itemize}
    \item \texttt{idx=0} $\rightarrow (1, 5)$
    \item \texttt{idx=1} $\rightarrow (2, 4)$
    \item \texttt{idx=2} $\rightarrow (3, 3)$
\end{itemize}

\[
\texttt{B} = [(1, 5, 0), (2, 4, 1), (3, 3, 2)]
\]

Sort \texttt{B} by $x$, then $y$, then \texttt{idx}:

\[
(1, 5, 0), \quad (2, 4, 1), \quad (3, 3, 2)
\]

---

\subsubsection*{Step 5: Setup \texttt{C} (BIT array) with $n = 5$}

Initialize 

\[
\texttt{C} = [0, 0, 0, 0, 0, 0] \quad \text{(1-based index: index 0 is unused)}
\]

---

\subsubsection*{Step 6: Initialize \texttt{D} and \texttt{j}}

\begin{itemize}
    \item \texttt{D} = [0, 0, 0] (same length as \texttt{B})
    \item \texttt{j} = 0
\end{itemize}

Now process each element in \texttt{B} in sorted order.

---

\textbf{Processing \texttt{B[0]} = (1, 5, 0)}

\textbf{Phase 1: Process \texttt{A2} while $j < n$ and $\texttt{A2}[j][0] < x$}
\begin{itemize}
    \item $x = 1$, compare \texttt{A2[j=0].0} = 0 $< 1$ $\rightarrow$ break
    \item No movement of $j$ $\rightarrow$ no calls to \texttt{upd()} (\red{Missing Logic Elements (code line 33 34 35): The model omits the entire loop.})
\end{itemize}

\textbf{Phase 2: Query}
\begin{itemize}
    \item $y = 5$, $x-1 = 0$
    \item Query $\texttt{qry}(5)$, then $\texttt{qry}(0)$. The result is $\texttt{qry}(5) - \texttt{qry}(0)$
\end{itemize}

\textbf{BIT: query(5)}

\begin{itemize}
    \item $i = 5 \rightarrow C[i] = 0, \quad i = 5 - 4 = 1 \rightarrow C[1] = 0$
    \item $i = 0 \rightarrow$ stop $\Rightarrow \texttt{qry}(5) = 0$
    \item $\texttt{bit\_ops} += 1$
\end{itemize}

\textbf{BIT: query(0)} returns 0 (base case), $\texttt{bit\_ops} += 1$

\[
\texttt{qry}(5) - \texttt{qry}(0) = 0 - 0 = 0
\]

\[
D[0] = 0, \quad \texttt{bit\_ops} = 2
\]

---

\paragraph*{Processing \texttt{B[1]} = (2, 4, 1)}

\textbf{Phase 1: Process \texttt{A2} while $j < n$ and $\texttt{A2}[j][0] < x=2$}

\begin{itemize}
    \item $\texttt{A2}[j=0].0 = 0 < 2$ $\rightarrow$ process \texttt{A2}[0] = (0, 1)
    
    \item Call \texttt{upd(1)}:
    \begin{itemize}
        \item $i = 1$
        \item Low bit of 1 = 1 $\Rightarrow C[1] += 1 \rightarrow C = [0, 1, 0, 0, 0, 0]$
        \item $i = 1 + 1 = 2$, low bit of 2 = 2 $\Rightarrow C[2] += 1$
        \item $i = 2 + 2 = 4$, low bit of 4 = 4 $\Rightarrow C[4] += 1$
        \item $i = 4 + 4 = 8$ $\rightarrow$ exit loop
        \item \texttt{bit\_ops} $+= 1$
    \end{itemize}
    
    \item $j = 1$
\end{itemize}

\begin{itemize}
    \item \texttt{A2}[j=1].0 = 0 $< 2$ $\rightarrow$ process (0, 2), call \texttt{upd(2)}:
    
    \begin{itemize}
        \item $C = [0, 1, 1, 0, 1, 0]$
        \item $i = 2 \rightarrow +2 \rightarrow i=4 \rightarrow +4 \rightarrow i=8 \rightarrow$ done
        \item Count incremented at positions 2, 4
        \item \texttt{bit\_ops} $+= 1$
    \end{itemize}
    
    \item $j = 2$
\end{itemize}

\begin{itemize}
    \item \texttt{A2}[j=2].0 = 0 $< 2$ $\rightarrow$ process (0, 3), call \texttt{upd(3)}:
    
    \begin{itemize}
        \item $C = [0, 1, 1, 1, 1, 0]$
        \item $i=3 \rightarrow +3 \rightarrow +4 \rightarrow i=7$ $\rightarrow$ stop at $i=7$
        \item \texttt{bit\_ops} $+= 1$
    \end{itemize}
    
    \item $j = 3$
\end{itemize}

\begin{itemize}
    \item \texttt{A2}[j=3].0 = 0 $< 2$ $\rightarrow$ process (0,4), call \texttt{upd(4)}:
    
    \begin{itemize}
        \item $C = [0, 1, 1, 1, 2, 0]$
        \item $i=4 \rightarrow +4 \rightarrow i=8$
        \item \texttt{bit\_ops} $+= 1$
    \end{itemize}
    
    \item $j = 4$
\end{itemize}

\begin{itemize}
    \item \texttt{A2}[j=4].0 = 0 $< 2$ $\rightarrow$ process (0,5), call \texttt{upd(5)}:
    
    \begin{itemize}
        \item $C = [0, 1, 1, 1, 2, 1]$
        \item $i=5 \rightarrow +1 \rightarrow i=6$ $\rightarrow 0$ end bit $\rightarrow i=6$
        \item \texttt{bit\_ops} $+= 1$
    \end{itemize}
    
    \item $j = 5$ $\rightarrow$ end loop
\end{itemize}

\textbf{Phase 2: Query}
\begin{itemize}
    \item $y = 4$, $x-1 = 1$
\end{itemize}

\textbf{BIT: query(4)}
\begin{itemize}
    \item $i=4 \rightarrow C[4] = 2$
    \item $i=0 \rightarrow$ stop? No $\rightarrow i = 4 - 4 = 0$
    \item Total: 2 $\rightarrow \texttt{bit\_ops} += 1$
\end{itemize}

\textbf{BIT: query(1)}
\begin{itemize}
    \item $i=1 \rightarrow C[1] = 1$
    \item $i = 1 - 1 = 0$ $\rightarrow$ done
    \item Total = 1 $\rightarrow \texttt{bit\_ops} += 1$
\end{itemize}

\[
\texttt{result} = 2 - 1 = 1
\]

\[
D[1] = 1, \quad \texttt{bit\_ops} = 7
\]

---

\paragraph*{Processing \texttt{B[2]} = (3, 3, 2)}

\textbf{Phase 1: Process \texttt{A2} while $j < n$ and $\texttt{A2}[j][0] < x = 3$}

Currently $j=5$ (no more to process)

\textbf{Phase 2: Query}

\begin{itemize}
    \item $y = 3$, $x-1 = 2$
\end{itemize}

\textbf{BIT: query(3)}

\begin{itemize}
    \item $i=3 \rightarrow C[3] = 1 \rightarrow i=2$
    \item $i=2 \rightarrow C[2] = 1 \rightarrow i=0 \rightarrow$ done
    \item Total = $1 + 1 = 2$ $\rightarrow \texttt{bit\_ops} += 1$
\end{itemize}

\textbf{BIT: query(2)}

\begin{itemize}
    \item $i=2 \rightarrow C[2] = 1 \rightarrow i=0 \rightarrow$ done
    \item Total = 1 $\rightarrow \texttt{bit\_ops} += 1$
\end{itemize}

\[
\texttt{result} = 2 - 1 = 1
\]

\[
D[2] = 1, \quad \texttt{bit\_ops} = 9
\]

---

\subsubsection*{Final \texttt{D} after all queries:}
\[
D = [0, 1, 1]
\]

---

\textbf{Statistics:}

\begin{itemize}
    \item \texttt{pop\_cnt = 4} (from initial stack operation)
    \item \texttt{bit\_ops = 9} (tracked step-by-step)
\end{itemize}

---

\textbf{Output:} \\
\textbf{Final result:} $[0, 1, 1]$

\textbf{Statistics:}  
\begin{lstlisting}[language=python]
{"stack_pops": 4, "bit_ops": 9}
\end{lstlisting}

---

\textbf{Answer:}

\textbf{Reasoning:} Step-by-step executed with all variables, logs, stack operations, and function calls to BIT including detailed motion and count of updates and queries.

\textbf{Output:} $[0, 1, 1]$

\textbf{Statistics:} \texttt{\{"stack\_pops": 4, "bit\_ops": 9\}}
\end{tcolorbox}
\clearpage
\subsection{Error Case: CodeForces-1481D}

\textbf{Error Type:} State Tracking Error

\textbf{Test Input:}
\begin{lstlisting}[breaklines=true,basicstyle=\scriptsize\ttfamily]
6
\end{lstlisting}

\textbf{LLM Results:}
\begin{lstlisting}[breaklines=true,basicstyle=\scriptsize\ttfamily]
output: [3, 4, 1, 5, 2, 6]
stats: {'heap_pushes': 6, 'max_heap_size': 3}
\end{lstlisting}

\textbf{Code Results:}
\begin{lstlisting}[breaklines=true,basicstyle=\scriptsize\ttfamily]
output: [3, 4, 1, 5, 2, 6]
stats: {'heap_pushes': 6, 'max_heap_size': 2}
\end{lstlisting}

\textbf{Function:}
\begin{lstlisting}[language=Python,breaklines=true,basicstyle=\scriptsize\ttfamily]
def f(n):
    import heapq
    A = [0] * n
    heap = []
    heapq.heappush(heap, (-n, 0, n - 1))
    heap_pushes = 1
    max_heap_size = len(heap)
    val = 1
    while heap:
        neg_len, x, y = heapq.heappop(heap)
        if len(heap) > max_heap_size:
            max_heap_size = len(heap)
        z = (x + y) // 2
        A[z] = val
        val += 1
        if x <= z - 1:
            slen = (z - 1) - x + 1
            heapq.heappush(heap, (-slen, x, z - 1))
            heap_pushes += 1
        if z + 1 <= y:
            slen = y - (z + 1) + 1
            heapq.heappush(heap, (-slen, z + 1, y))
            heap_pushes += 1
    return A, {"heap_pushes": heap_pushes, "max_heap_size": max_heap_size}
\end{lstlisting}

\begin{tcolorbox}[
    colback=blue!5!white,
    colframe=blue!25!black,
    title=Function Description,
    fonttitle=\bfseries,
    fontupper=\scriptsize,
    rounded corners,
    drop shadow,
    breakable,
    left=3pt,
    right=3pt
]
\textbf{INPUTS:} One input: an integer \texttt{n}. From that you're going to make a list called \texttt{A} of length \texttt{n} filled with zeros. You'll also use a list called \texttt{heap} to store triples of numbers, and keep three counters: \texttt{heap\_pushes}, \texttt{max\_heap\_size}, and \texttt{val}.\\

\textbf{LOGICS:} Start by creating a list called \texttt{A} with \texttt{n} zeros in it, and an empty list called \texttt{heap}. Then push the triple $(-n, 0, n-1)$ onto \texttt{heap}, set \texttt{heap\_pushes} to 1, set \texttt{max\_heap\_size} to 1, and set a counter \texttt{val} to 1. Now you're going to repeat these steps as long as \texttt{heap} is not empty.\\
First, remove from \texttt{heap} the triple that has the smallest first number; if there's a tie, pick the one with the smallest second number. Call the three parts \texttt{neg\_len}, \texttt{x}, and \texttt{y}. Right after you take it out, check the current size of \texttt{heap}; if it's bigger than \texttt{max\_heap\_size}, update \texttt{max\_heap\_size} to that size.\\
Next, figure out \texttt{z} by adding \texttt{x} and \texttt{y}, dividing by 2 with integer division (discard any remainder). Then put \texttt{val} into \texttt{A} at index \texttt{z}, and add 1 to \texttt{val}.\\
What you do next is see if a left piece remains: check if \texttt{x} is less than or equal to \texttt{z} $-$ 1, and if so compute \texttt{slen} as $(z-1) - x + 1$, push the triple $(-\texttt{slen}, x, z-1)$ onto \texttt{heap}, and add 1 to \texttt{heap\_pushes}.\\
Then check if a right piece remains: see if \texttt{z} $+$ 1 is less than or equal to \texttt{y}, and if it does compute \texttt{slen} as $y - (z+1) + 1$, push the triple $(-\texttt{slen}, z+1, y)$ onto \texttt{heap}, and add 1 to \texttt{heap\_pushes}.\\
After you do both, go back to the start of these steps. When \texttt{heap} is empty, the loop ends.\\

\textbf{OUTPUTS:} At the end, you'll have your list \texttt{A} filled with numbers from 1 up to \texttt{n} in a certain order, and two counters: \texttt{heap\_pushes}, which records how many times you pushed a triple onto \texttt{heap}, and \texttt{max\_heap\_size}, which tells you the largest number of triples that ever sat in \texttt{heap} at once.
\end{tcolorbox}

\begin{tcolorbox}[
    colback=gray!3!white,
    colframe=gray!30!black,
    title=LLM Reasoning,
    fonttitle=\bfseries,
    fontupper=\scriptsize,
    rounded corners,
    drop shadow,
    breakable
]
\textbf{Reasoning:}

\vspace{0.3cm}

\textbf{Given input}: \texttt{n = 6}

\vspace{0.3cm}

\textbf{Step 1: Initialization}

\begin{itemize}
    \item Create \texttt{A = [0, 0, 0, 0, 0, 0]}
    \item \texttt{heap = []} (empty)
    \item Push \texttt{(-6, 0, 5)} onto heap\\
    (heap now: \texttt{[(-6, 0, 5)]})
    \item \texttt{heap\_pushes = 1}
    \item \texttt{max\_heap\_size = 1}
    \item \texttt{val = 1}
\end{itemize}

\noindent\rule{\linewidth}{0.4pt}

\vspace{0.3cm}

\textbf{Step 2: Main loop (while heap not empty)}

\vspace{0.3cm}

\textbf{Iteration 1}
\begin{itemize}
    \item heap: \texttt{[(-6, 0, 5)]}
    \item Pop \texttt{(-6, 0, 5)}. (heap is now empty)
    \begin{itemize}
        \item After pop, heap size: 0 
        \item max\_heap\_size stays 1
    \end{itemize}
    \item \texttt{neg\_len = -6, x = 0, y = 5}
    \item \texttt{z = (0 + 5) // 2 = 2}
    \item Set \texttt{A[2] = 1} \(\to\) \texttt{A = [0, 0, 1, 0, 0, 0]}
    \item \texttt{val = 2}
    \item Left piece: \texttt{x=0, z-1=1} so \(0 \leq 1\)
    \begin{itemize}
        \item \texttt{slen = (1) - 0 + 1 = 2}
        \item Push \texttt{(-2, 0, 1)} onto heap (\texttt{heap\_pushes=2})
        \item heap: \texttt{[(-2, 0, 1)]}
    \end{itemize}
    \item Right piece: \texttt{z+1=3, y=5} so \(3 \leq 5\)
    \begin{itemize}
        \item \texttt{slen = 5-(3)+1 = 3}
        \item Push \texttt{(-3, 3, 5)} onto heap (\texttt{heap\_pushes=3})
        \item heap: \texttt{[(-3, 3, 5), (-2, 0, 1)]}
    \end{itemize}
    \item max\_heap\_size now 2
\end{itemize}

\noindent\rule{\linewidth}{0.4pt}

\vspace{0.3cm}

\textbf{Iteration 2}
\begin{itemize}
    \item heap: \texttt{[(-3, 3, 5), (-2, 0, 1)]}
    \item Pop \texttt{(-3, 3, 5)}
    \begin{itemize}
        \item After pop, heap size: 1
        \item max\_heap\_size stays 2
    \end{itemize}
    \item \texttt{neg\_len=-3, x=3, y=5}
    \item \texttt{z = (3+5)//2 = 4}
    \item Set \texttt{A[4] = 2} \(\to\) \texttt{A = [0, 0, 1, 0, 2, 0]}
    \item \texttt{val = 3}
    \item Left piece: \texttt{x=3, z-1=3}, so \(3 \leq 3\)
    \begin{itemize}
        \item \texttt{slen = (3)-(3)+1=1}
        \item Push \texttt{(-1, 3, 3)} (\texttt{heap\_pushes=4})
        \item heap: \texttt{[(-2, 0, 1), (-1, 3, 3)]}
    \end{itemize}
    \item Right piece: \texttt{z+1=5, y=5} so \(5 \leq 5\)
    \begin{itemize}
        \item \texttt{slen = 5-(5)+1=1}
        \item Push \texttt{(-1, 5, 5)} (\texttt{heap\_pushes=5})
        \item heap: \texttt{[(-2, 0, 1), (-1, 3, 3), (-1, 5, 5)]}
    \end{itemize}
    \item max\_heap\_size now 3 (\red{State Tracking Error: The model should not update \texttt{max\_heap\_size} here.})
\end{itemize}

\noindent\rule{\linewidth}{0.4pt}

\vspace{0.3cm}

\textbf{Iteration 3}
\begin{itemize}
    \item heap: \texttt{[(-2, 0, 1), (-1, 3, 3), (-1, 5, 5)]}
    \item Pop \texttt{(-2, 0, 1)}
    \begin{itemize}
        \item After pop, heap size: 2
        \item max\_heap\_size stays 3
    \end{itemize}
    \item \texttt{neg\_len=-2, x=0, y=1}
    \item \texttt{z = (0+1)//2 = 0}
    \item Set \texttt{A[0] = 3} \(\to\) \texttt{[3, 0, 1, 0, 2, 0]}
    \item \texttt{val = 4}
    \item Left piece: \texttt{x=0, z-1=-1} --- \(0 \leq -1\) is False, no left piece
    \item Right piece: \texttt{z+1=1, y=1} --- \(1 \leq 1\) is True
    \begin{itemize}
        \item \texttt{slen = 1-1+1=1}
        \item Push \texttt{(-1, 1, 1)} (\texttt{heap\_pushes=6})
        \item heap: \texttt{[(-1, 1, 1), (-1, 3, 3), (-1, 5, 5)]}
    \end{itemize}
    \item max\_heap\_size stays 3
\end{itemize}

\noindent\rule{\linewidth}{0.4pt}

\vspace{0.3cm}

\textbf{Iteration 4}
\begin{itemize}
    \item heap: \texttt{[(-1, 1, 1), (-1, 3, 3), (-1, 5, 5)]}
    \item Pop \texttt{(-1, 1, 1)}
    \begin{itemize}
        \item After pop, heap size: 2
    \end{itemize}
    \item \texttt{neg\_len=-1, x=1, y=1}
    \item \texttt{z = (1+1)//2 = 1}
    \item Set \texttt{A[1] = 4} \(\to\) \texttt{[3, 4, 1, 0, 2, 0]}
    \item \texttt{val = 5}
    \item Left piece: \(1 \leq 0\)? False
    \item Right piece: \(2 \leq 1\)? False
    \item No pushes
\end{itemize}

\noindent\rule{\linewidth}{0.4pt}

\vspace{0.3cm}

\textbf{Iteration 5}
\begin{itemize}
    \item heap: \texttt{[(-1, 3, 3), (-1, 5, 5)]}
    \item Pop \texttt{(-1, 3, 3)}
    \begin{itemize}
        \item After pop, heap size: 1
    \end{itemize}
    \item \texttt{neg\_len=-1, x=3, y=3}
    \item \texttt{z = (3+3)//2 = 3}
    \item Set \texttt{A[3] = 5} \(\to\) \texttt{[3, 4, 1, 5, 2, 0]}
    \item \texttt{val = 6}
    \item Left piece: \(3 \leq 2\)? False
    \item Right piece: \(4 \leq 3\)? False
    \item No pushes
\end{itemize}

\noindent\rule{\linewidth}{0.4pt}

\vspace{0.3cm}

\textbf{Iteration 6}
\begin{itemize}
    \item heap: \texttt{[(-1, 5, 5)]}
    \item Pop \texttt{(-1, 5, 5)}
    \begin{itemize}
        \item After pop, heap size: 0
    \end{itemize}
    \item \texttt{neg\_len=-1, x=5, y=5}
    \item \texttt{z = (5+5)//2 = 5}
    \item Set \texttt{A[5] = 6} \(\to\) \texttt{[3, 4, 1, 5, 2, 6]}
    \item \texttt{val = 7}
    \item Left piece: \(5 \leq 4\)? False
    \item Right piece: \(6 \leq 5\)? False
    \item No pushes
\end{itemize}

\noindent\rule{\linewidth}{0.4pt}

\vspace{0.3cm}

HEAP IS NOW EMPTY

\vspace{0.3cm}
\noindent\rule{\linewidth}{0.4pt}

\vspace{0.3cm}

\textbf{Final Results:}

\begin{itemize}
    \item \textbf{A}: \texttt{[3, 4, 1, 5, 2, 6]}
    \item \textbf{heap\_pushes}: 6
    \item \textbf{max\_heap\_size}: 3
\end{itemize}

\vspace{0.3cm}

\noindent\rule{\linewidth}{0.4pt}

\vspace{0.3cm}

\textbf{Output:}\\
\texttt{[3, 4, 1, 5, 2, 6]}

\vspace{0.3cm}

\textbf{Statistics:}\\
\texttt{\{'heap\_pushes': 6, 'max\_heap\_size': 3\}}
\end{tcolorbox}

\clearpage
\section{Error Rate vs Logic Complexity}\label{sec:error_vs_logic}
\begin{figure*}[h]
    \centering
    % First row
    \begin{minipage}[t]{0.32\textwidth}
        \centering
        \includegraphics[width=\textwidth]{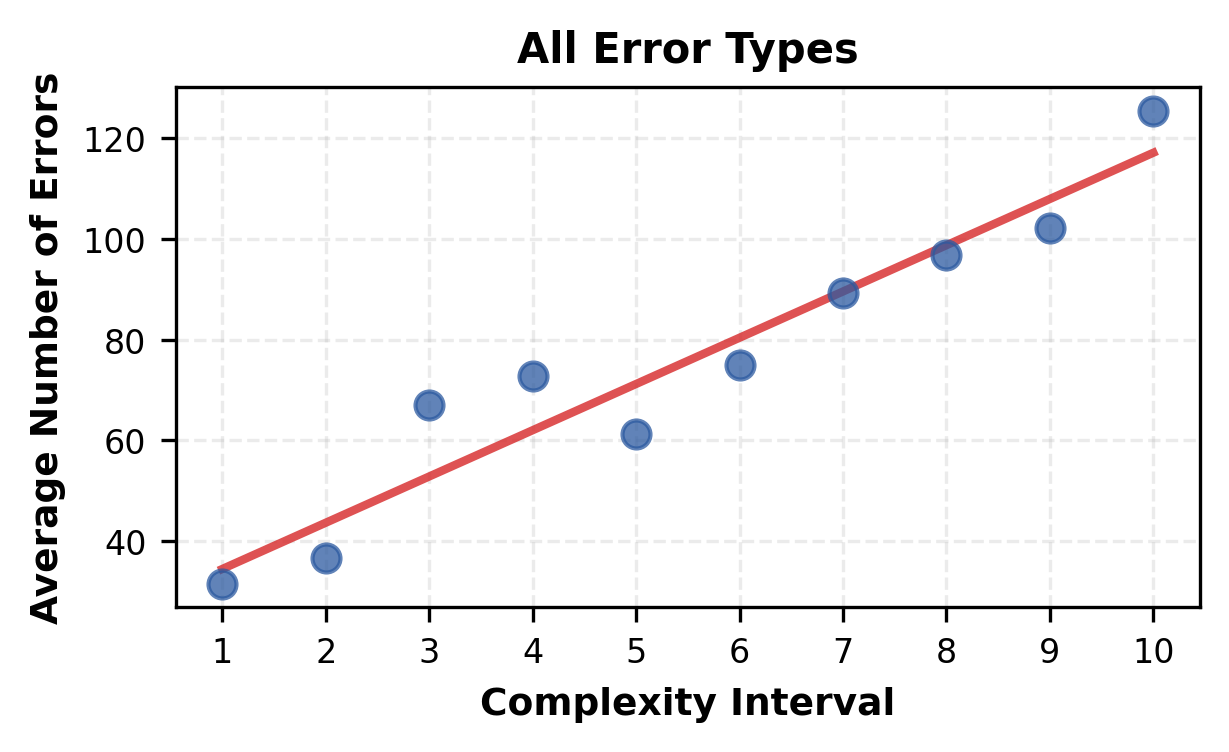}
    \end{minipage}%
    \hfill
    \begin{minipage}[t]{0.32\textwidth}
        \centering
        \includegraphics[width=\textwidth]{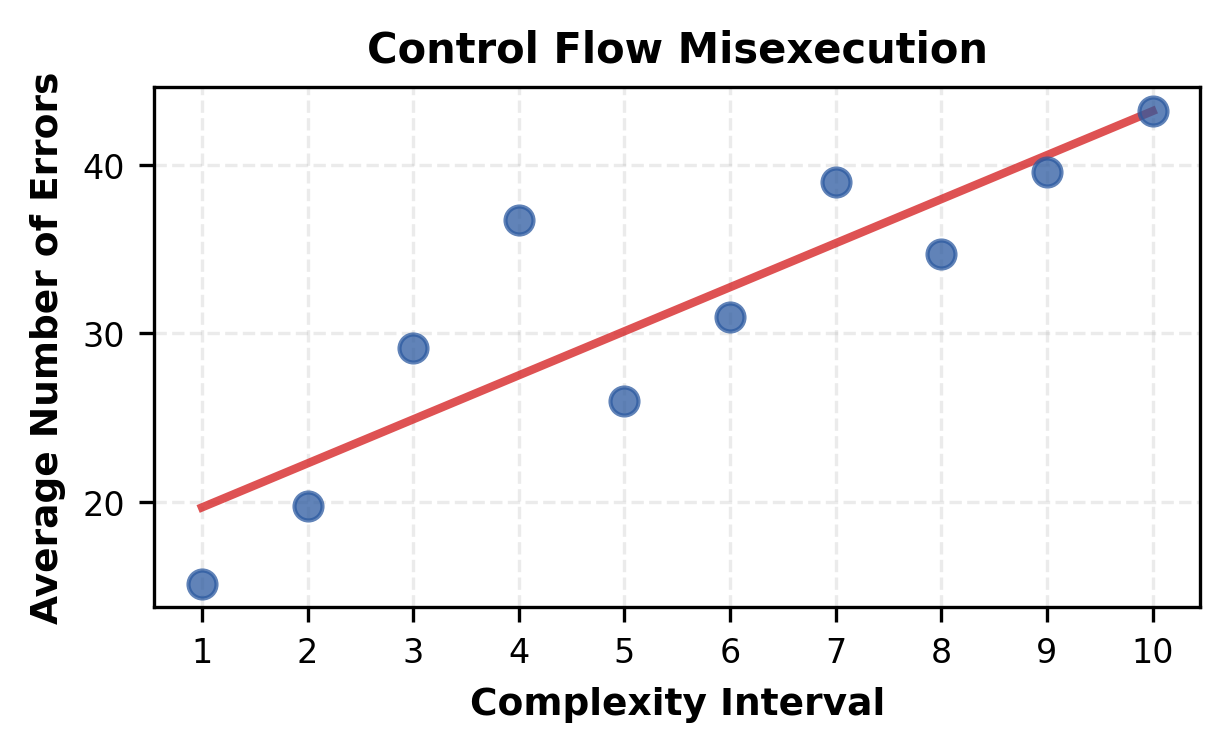}
    \end{minipage}%
    \hfill
    \begin{minipage}[t]{0.32\textwidth}
        \centering
        \includegraphics[width=\textwidth]{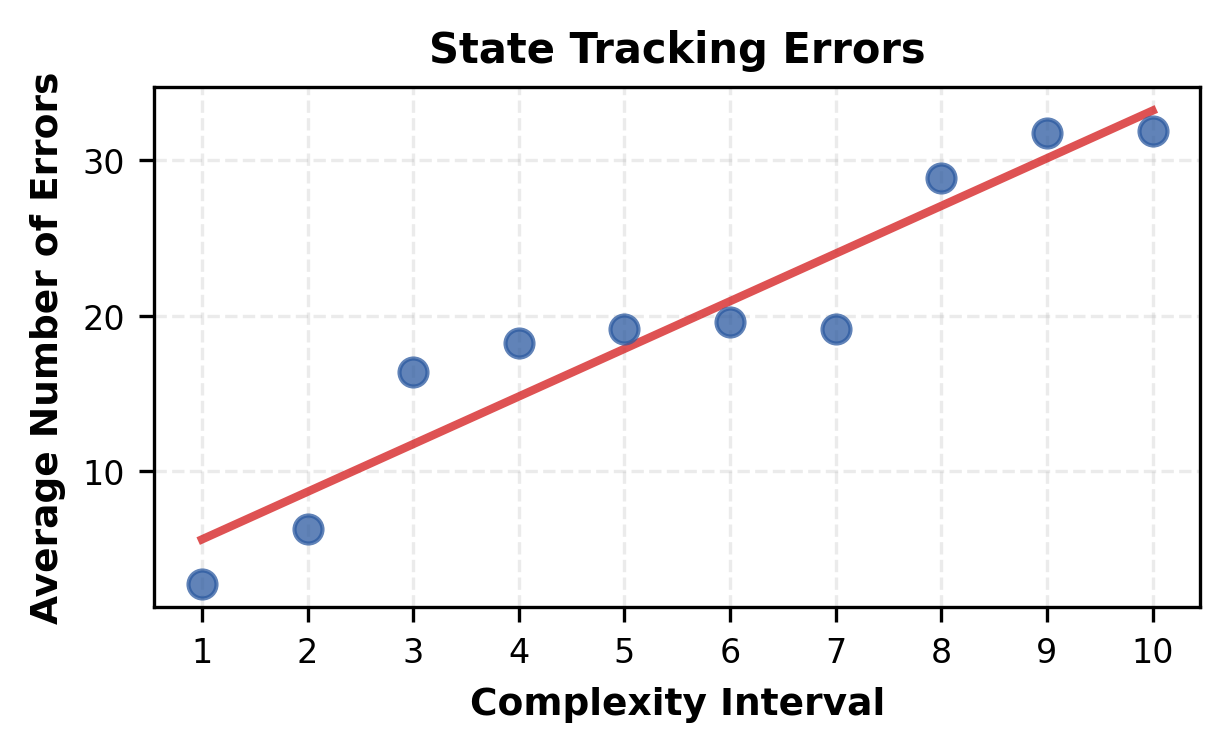}
    \end{minipage}
    
    % \vspace{0.3cm}
    
    % Second row
    \begin{minipage}[t]{0.32\textwidth}
        \centering
        \includegraphics[width=\textwidth]{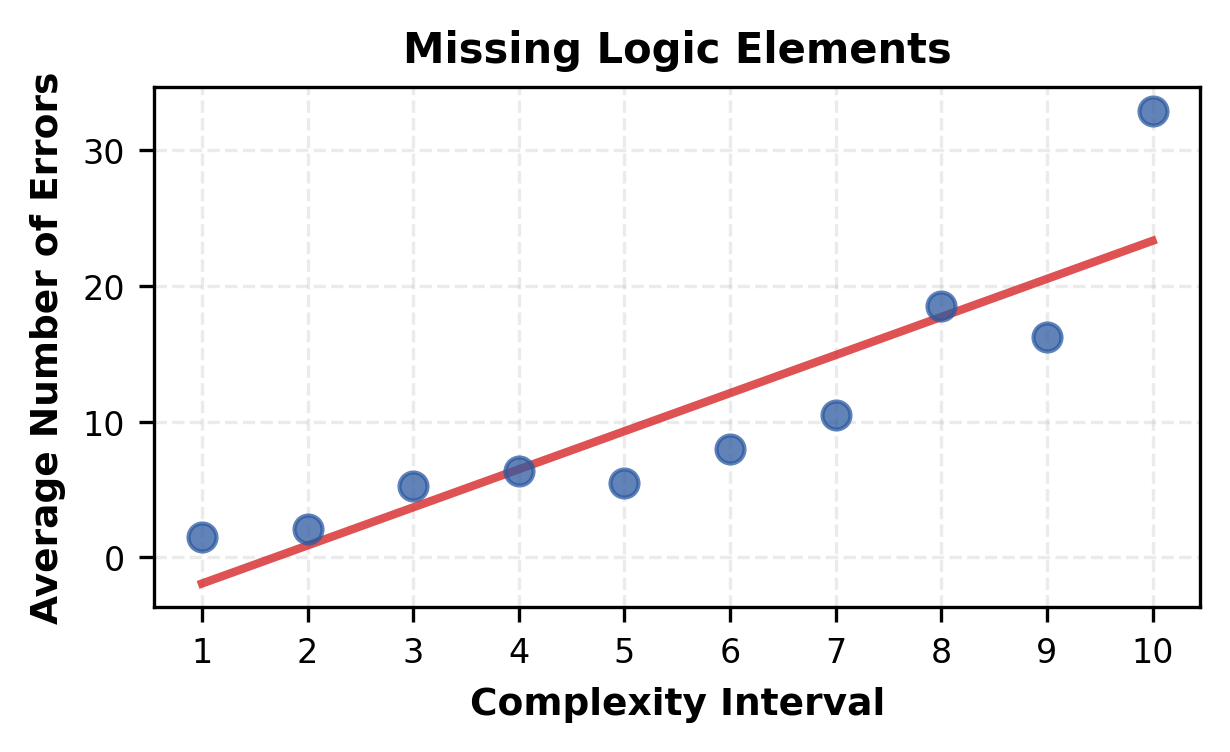}
    \end{minipage}%
    \hfill
    \begin{minipage}[t]{0.32\textwidth}
        \centering
        \includegraphics[width=\textwidth]{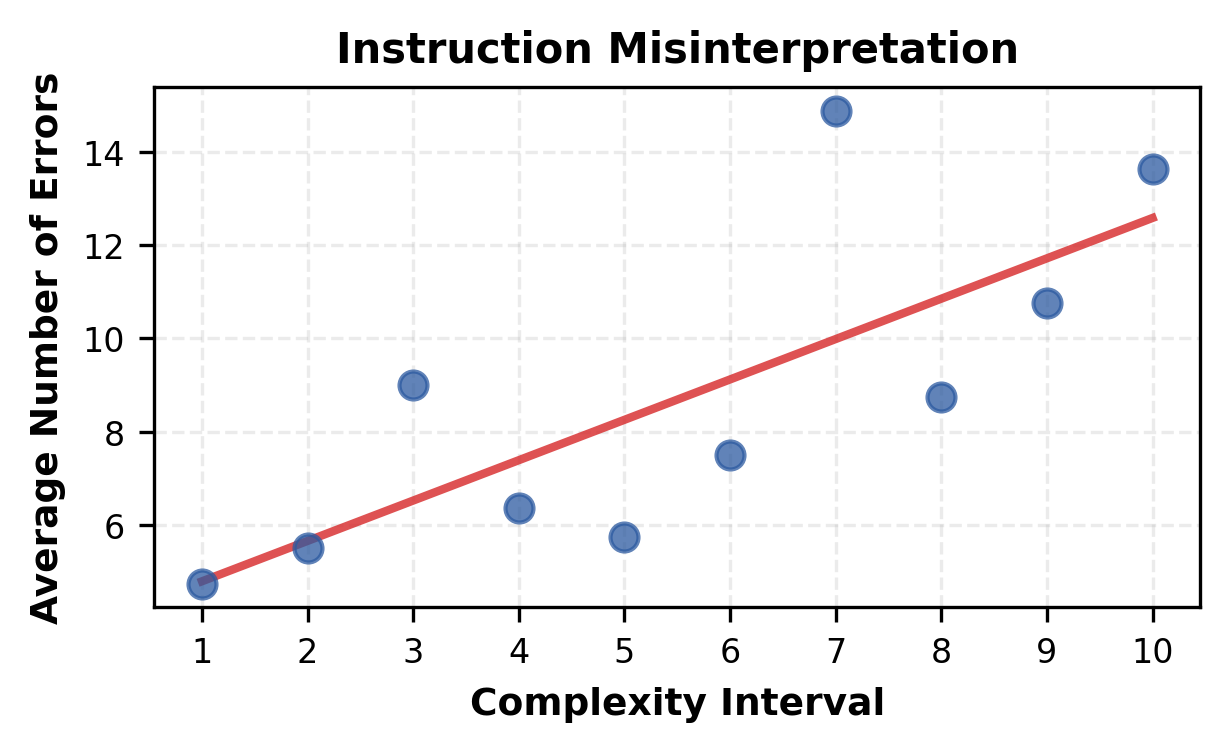}
    \end{minipage}%
    \hfill
    \begin{minipage}[t]{0.32\textwidth}
        \centering
        \includegraphics[width=\textwidth]{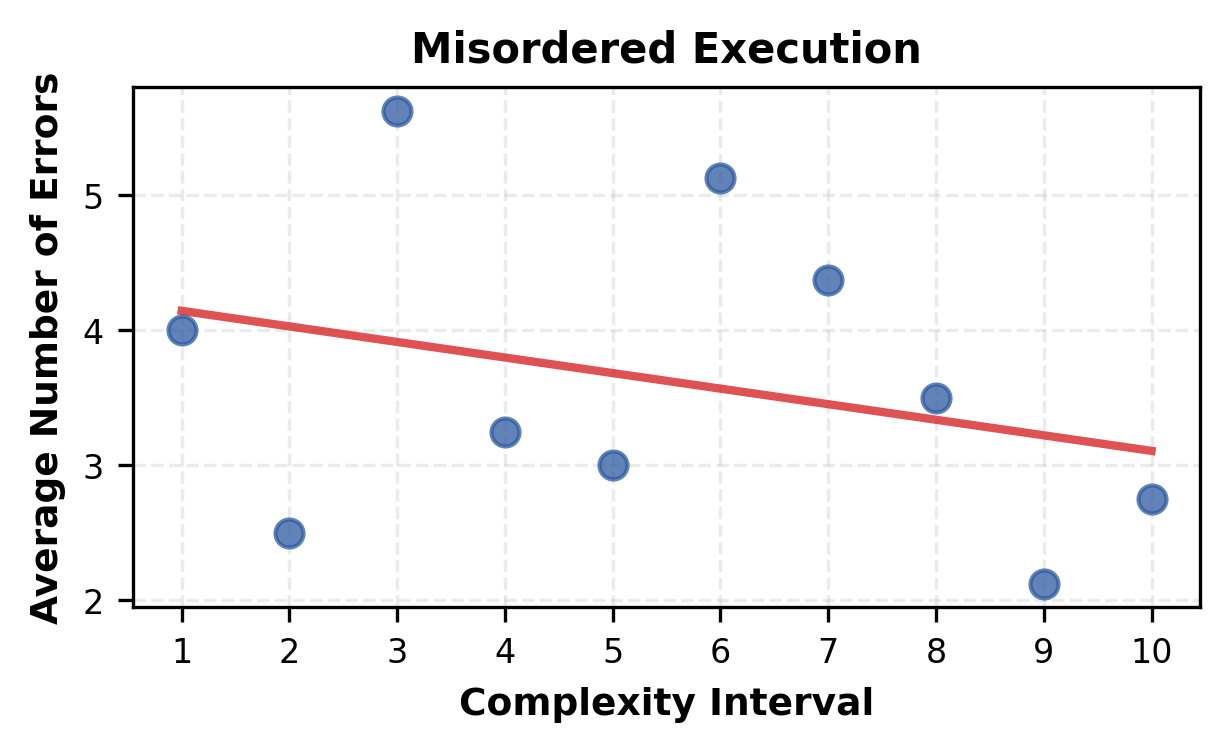}
    \end{minipage}
    
    \caption{Error distribution across complexity intervals. Blue points represent average error counts across 8 models at each complexity interval, with red lines indicating linear trends.}
    \label{fig:error_trend}
\end{figure*}

To analyze the relationship between error rate and logical complexity, we divide the instructions in \bench{} into ten intervals based on their complexity scores computed using Eq.~\ref{eq:score}. Interval 1 corresponds to the easiest instructions, while Interval 10 contains the most difficult ones. For each interval, we compute the average number of error cases across all eight representative models and visualize the trend in Figure~\ref{fig:error_trend}. \textbf{The results show a clear positive correlation between logical complexity and error frequency: as complexity increases, models produce substantially more errors}. In particular, the proportions of Control Flow Misexecution, Instruction Misinterpretation, Missing Logic Elements, and State Tracking Errors rise sharply, suggesting that current LLMs mainly struggle with understanding and executing complex logical structures. In contrast, Misordered Execution remains consistently low across all complexity levels, indicating that models generally follow the execution order specified in the instructions.

% The top-left panel shows all error types combined, while the other panels show individual failure modes.

\clearpage
\section{Prompts}

\begin{figure}[htbp]
\centering
\begin{minipage}{0.95\textwidth}
\footnotesize
\begin{lstlisting}[frame=single, breaklines=true, basicstyle=\ttfamily\scriptsize, numbers=none]
You are an expert Python programmer. I want to enhance a function by adding meaningful, NON-REDUNDANT execution statistics that capture the most crucial aspects of its computational behavior.

**CRITICAL REQUIREMENTS:**
1. **Adaptive number of statistics (1-3)** - Choose the appropriate number based on function complexity
2. **No redundant statistics** - Each must measure something completely different
3. **Focus on crucial logic** - Statistics should reflect key computational steps that matter
4. **Generic variable names** - Convert all meaningful variable names to generic ones

**Variable Naming Guidelines:**
- Lists/arrays: Use "L", "arr", "lst", "A", "B" etc.
- Matrices/2D arrays: Use "M", "matrix", "grid" etc.
- Strings: Use "s", "text", "str1", "str2" etc.
- Integers/numbers: Use "n", "m", "x", "y", "val" etc.
- Dictionaries/maps: Use "d", "mp", "cache" etc.
- Sets: Use "st", "visited", "seen" etc.
- Keep parameter names generic but clear about data types

**Guidelines for choosing the number of statistics:**
- **1 statistic**: Simple functions with one main operation (e.g., single loop, basic calculation)
- **2 statistics**: Functions with two distinct computational aspects (e.g., nested structure with two key operations)
- **3 statistics**: Complex functions with multiple distinct computational phases or operations

**What makes a good statistic:**
- Counts operations that directly affect the algorithm's behavior
- Measures key computational steps that vary with different inputs
- Reflects important decision points or iterations in the logic
- Avoids counting trivial operations (simple assignments, basic comparisons)

Original Function:
```python
{function}
```

**Requirements:**
1. Convert all meaningful variable names to generic ones (following naming guidelines above)
2. Change the function name to 'f' (regardless of original name)
3. Remove ALL comments from the function code
4. Return a tuple: (original_output, statistics_dict)
5. Choose 1-3 statistics based on the function's complexity and distinct computational aspects
6. Each statistic measures a unique computational aspect
7. Focus on operations that directly affect the function's core logic
8. Preserve the exact same logic and functionality, only change variable names and remove comments

Return your response in JSON format:
{{
    "function": "complete modified function code here with generic variable names, function name 'f', and no comments",
    "stats_keys": ["list", "of", "1", "to", "3", "unique", "keys"]
}}
\end{lstlisting}
\end{minipage}
\caption{Prompt for Adding State Trackers and Anonymizing Functions}
\label{fig:prompt_add_stat}
\end{figure}

\begin{figure}[htbp]
\centering
\begin{minipage}{0.95\textwidth}
\footnotesize
\begin{lstlisting}[frame=single, breaklines=true, basicstyle=\ttfamily\scriptsize, numbers=none]
You are helping someone understand how to manually process data step-by-step. I'll give you a function, and I need you to explain it like you're talking to someone who needs to do this work by hand - as if you're giving them verbal instructions over the phone. Your explanation should be so clear and detailed that they can follow along exactly and get the same results.

**CRITICAL REQUIREMENT**: Your instructions must be so complete and precise that someone could follow them step-by-step WITHOUT seeing any code and produce the exact same output and statistics as the function would.

**IMPORTANT CONSTRAINTS:**
- Don't explain what this is used for in the real world - just focus on the data processing steps
- Don't mention specific problem names or applications
- Treat this as pure data manipulation work
- Use the generic variable names from the code (L, arr, n, m, etc.)

**Natural Language Guidelines:**
- Write like you're speaking to someone conversationally
- Use natural transitions like "Now you need to...", "Next, go through...", "At this point..."
- Include phrases like "What you're going to do is...", "The way this works is..."
- Make it sound like verbal instructions, not a formal manual
- Still be extremely precise about every detail, but use conversational language
- Use connecting words and phrases that make it flow naturally
- Include every conditional check, loop, and decision point in natural speech
- Be specific about indexing, bounds, and conditions, but explain them conversationally

Function to describe:
```python
{function}
```

Return your response in the following JSON format with exactly three sections:
{{
    "inputs": "Describe the data types and structure of what you're working with (like 'a list of numbers' or 'two text strings'). Don't mention what these represent in real-world terms.",
    "logics": "Give detailed, conversational instructions for processing the data step-by-step. Use natural language like you're talking someone through it, with phrases like 'Now you need to...', 'Next, go through...', 'What you do is...'. Be extremely precise about every step, condition, and operation, but explain it in a flowing, conversational way. Include all loops, decisions, calculations, and counter updates. Use the generic variable names from the code.",
    "outputs": "Explain what you'll end up with and what each number in your statistics dictionary represents from the work you did."
}}
\end{lstlisting}
\end{minipage}
\caption{Prompt for Generating Natural Language Description}
\label{fig:prompt_gen_desc}
\end{figure}

\begin{figure}[htbp]
\centering
\begin{minipage}{0.95\textwidth}
\footnotesize
\begin{lstlisting}[frame=single, breaklines=true, basicstyle=\ttfamily\scriptsize, numbers=none]
You are helping someone check if a set of conversational instructions are complete enough for manual data processing. I need you to verify whether these step-by-step instructions would allow someone to manually work through the data and get the same results as the code - like you're checking if verbal instructions over the phone would be complete enough for someone to follow along exactly.

**CRITICAL REQUIREMENT**: The instructions must be so complete and precise that someone could follow them step-by-step WITHOUT seeing any code and produce the exact same output and statistics as the function would.

**VERIFICATION APPROACH:**
Think about this like you're helping someone understand whether these conversational instructions are good enough for manual data processing. The instructions should be so clear and complete that someone could:
1. **Follow Every Step**: All the data processing steps are explained like you're talking someone through it
2. **Handle All Cases**: Every condition, loop, and decision point is covered in natural language
3. **Track Everything**: All variable updates, counters, and calculations are explained conversationally
4. **Get Same Results**: Following the instructions would produce identical output and statistics
5. **No Guessing**: Every significant operation is covered so no one has to guess what to do
6. **Natural Flow**: The instructions flow naturally like someone talking through the process

**FUNCTION CODE:**
```python
{function_code}
```

**CONVERSATIONAL INSTRUCTIONS TO CHECK:**
{description}

**COMPLETENESS JUDGMENT:**
- **COMPLETE**: The conversational instructions cover everything needed - someone could follow them and get identical results
- **INCOMPLETE**: Some operations, conditions, or steps are missing or unclear - following the instructions wouldn't match the code's behavior

Return your response in JSON format:
{{
    "desc_is_complete": true/false,
    "reasoning": "Talk through your assessment conversationally, like you're explaining to someone what's working well in these instructions and what might be missing. Use natural language like 'What I notice is...', 'The instructions do a good job of...', 'What's missing is...', 'Someone following these would probably get confused when...'",
    "missing_aspects": ["List specific operations or steps that aren't covered conversationally - describe them in natural language like 'explaining how to update the counter', 'walking through the loop condition', 'describing what to do when the list is empty'"],
    "coverage_percentage": "estimated percentage (0-100) of code operations covered by the conversational instructions"
}}


\end{lstlisting}
\end{minipage}
\caption{Prompt for Verifing Natural Language Description}
\label{fig:prompt_verify_desc}
\end{figure}

\begin{figure}[htbp]
\centering
\begin{minipage}{0.95\textwidth}
\footnotesize
\begin{lstlisting}[frame=single, breaklines=true, basicstyle=\ttfamily\scriptsize, numbers=none]
You are an expert Python programmer. I want you to evolve an existing function to make it MORE LOGICALLY COMPLICATED while maintaining the exact same input signature and core functionality. The evolved function should be significantly more sophisticated in its logic and computational approach.

**CRITICAL REQUIREMENTS:**
1. **Same Input Signature**: The function must accept exactly the same parameters as the original
2. **Same Core Output**: The main result should be equivalent to the original function's output
3. **More Complex Logic**: Add sophisticated algorithmic patterns, advanced data structures, or multi-phase processing
4. **Enhanced Statistics**: Statistics can change to reflect the new complexity (1-3 meaningful stats)
5. **Preserve Function Name**: Keep the function name as 'f'
6. **Return Format**: Must return tuple (result, stats_dict)
7. **ABSOLUTELY NO COMMENTS**: Do NOT write any comments, docstrings, or explanations in the evolved function code. The function must be completely comment-free.

**EVOLUTION STRATEGIES (choose the most appropriate):**
- **Multi-phase processing**: Break the problem into sophisticated stages
- **Advanced data structures**: Use heaps, trees, graphs, or complex mappings
- **Optimized algorithms**: Replace naive approaches with efficient algorithms
- **Dynamic programming**: Add memoization or tabulation for overlapping subproblems
- **Divide and conquer**: Split problem into smaller, more complex subproblems
- **State machines**: Add complex state tracking and transitions
- **Mathematical optimization**: Add advanced mathematical techniques
- **Sophisticated filtering/sorting**: Use multiple criteria or advanced comparison logic

**STATISTICS GUIDELINES:**
- Choose 1-3 statistics that reflect the NEW complexity
- Track operations that highlight the sophisticated logic
- Examples: phases_completed, recursive_calls, cache_hits, comparisons, transformations, iterations

**ORIGINAL FUNCTION:**
```python
{original_function}
```

**REQUIREMENTS:**
1. Analyze the original function's core purpose and constraints
2. Design a more sophisticated approach that achieves the same goal
3. Implement complex logic patterns while preserving correctness
4. Add meaningful statistics that capture the new complexity
5. Ensure the evolved function is significantly more algorithmically interesting
6. Test edge cases and maintain robustness
7. **ABSOLUTELY NO COMMENTS**: The evolved function code must be completely comment-free

**OUTPUT FORMAT:**
```json
{{
    "evolved_function": "def f(...):\n    stat1 = 0\n    stat2 = 0\n    \n    complex_logic_here\n    \n    return result, {{'stat1': stat1, 'stat2': stat2}}",
    "stats_keys": ["stat1", "stat2"],
    "evolution_description": "Brief explanation of how the function was made more complex (e.g., 'Added multi-phase processing with dynamic programming', 'Implemented graph-based approach with state tracking')"
}}
```
\end{lstlisting}
\end{minipage}
\caption{Prompt for Function Evolution}
\label{fig:prompt_evolve}
\end{figure}

\clearpage
\section{Human Verification}\label{sec:human_verify}
To ensure the quality of the generated instructions, we conducted a human verification study on 136 randomly sampled instances from \bench{}. The data were split evenly into four parts, each containing 34 instructions, and each part was assigned to two independent PhD-level annotators with expertise in computer science. Annotators were instructed to verify whether every line of the anonymized function was accurately and completely captured in the corresponding natural language instruction. For Batches~1 and~2, both annotators fully agreed on all items, yielding 100\% agreement. In Batch~3, both annotators marked 32 out of 34 items as correct, with one identical negative case, resulting in an agreement rate of $33/34 \,(97.06\%)$. In Batch~4, the two annotators agreed on 32 out of 34 items ($94.12\%$). Overall, the human agreement rate across all 136 items was $\frac{133}{136} \approx 97.79\%$, demonstrating a high level of consistency and validating the reliability of the generated instructions.

\section{Cases of Thinking Helps Large LLMs}\label{thinking_case}
% Error examples for paper
% Generated automatically from error analysis

\subsection{Case 1: CODEFORCES-818D}\label{think_case_1}

\textbf{Test Input:}
\begin{lstlisting}[breaklines=true,basicstyle=\scriptsize\ttfamily]
6, [3, 1, 4, 1, 5, 9] (len=6)
\end{lstlisting}

\textbf{LLM Results:}
\begin{lstlisting}[breaklines=true,basicstyle=\scriptsize\ttfamily]
output: 2
stats: {'left_pops': 5, 'right_pops': 2, 'balanced_subarrays': 19}
\end{lstlisting}

\textbf{Code Results:}
\begin{lstlisting}[breaklines=true,basicstyle=\scriptsize\ttfamily]
output: 2
stats: {'left_pops': 5, 'right_pops': 2, 'balanced_subarrays': 19}
\end{lstlisting}

\textbf{Function:}
\begin{lstlisting}[language=Python,breaklines=true,basicstyle=\scriptsize\ttfamily]
def f(n, A):
    prev = [-1] * n
    nxt = [n] * n
    st = []
    l_pops = 0
    for i in range(n):
        while st and A[st[-1]] <= A[i]:
            st.pop()
            l_pops += 1
        prev[i] = st[-1] if st else -1
        st.append(i)
    st.clear()
    r_pops = 0
    for i in range(n-1, -1, -1):
        while st and A[st[-1]] < A[i]:
            st.pop()
            r_pops += 1
        nxt[i] = st[-1] if st else n
        st.append(i)
    total = n * (n + 1) // 2
    balanced = 0
    for i in range(n):
        x = i - prev[i]
        y = nxt[i] - i
        balanced += x + y - 1
    return total - balanced, {'left_pops': l_pops, 'right_pops': r_pops, 'balanced_subarrays': balanced}
\end{lstlisting}

\begin{tcolorbox}[
    colback=gray!3!white,
    colframe=gray!30!black,
    title=LLM Thinking Process,
    fonttitle=\bfseries,
    fontupper=\scriptsize,
    rounded corners,
    drop shadow,
    breakable
]
Okay, I need to execute this algorithm step-by-step manually. Let me break down what's happening:

\textbf{INPUT:} \( n = 6 \), \( A = [3, 1, 4, 1, 5, 9] \)

\textbf{Step 1: Initialize data structures}
\begin{itemize}
    \item \texttt{prev} = \([-1, -1, -1, -1, -1, -1]\) (n slots, all \(-1\))
    \item \texttt{nxt} = \([6, 6, 6, 6, 6, 6]\) (n slots, all \(n\))
    \item \texttt{st} = \([]\) (empty stack)
    \item \texttt{l\_pops} = 0
\end{itemize}

\textbf{Step 2: Forward sweep (i from 0 to n-1)}\\
I need to go through each \(i\) from 0 to 5:

\textbf{\(i = 0\):}
\begin{itemize}
    \item \(A[0] = 3\)
    \item \texttt{st} is empty, so no popping
    \item \texttt{st} is empty, so \texttt{prev[0]} = \(-1\) (already is \(-1\))
    \item push 0 onto \texttt{st}: \(\texttt{st} = [0]\)
\end{itemize}

\textbf{\(i = 1\):}
\begin{itemize}
    \item \(A[1] = 1\)
    \item \(\texttt{st} = [0]\), top is 0, \(A[0] = 3\)
    \item \(A[0] = 3 > A[1] = 1\), so no popping (condition is \(A[\text{top}] \leq A[i]\))
    \item \texttt{st} is not empty, top is 0, so \(\texttt{prev[1]} = 0\)
    \item push 1 onto \texttt{st}: \(\texttt{st} = [0, 1]\)
\end{itemize}

\textbf{\(i = 2\):}
\begin{itemize}
    \item \(A[2] = 4\)
    \item \(\texttt{st} = [0, 1]\), top is 1, \(A[1] = 1\)
    \item \(A[1] = 1 \leq A[2] = 4\), so pop 1, \texttt{l\_pops} = 1, \(\texttt{st} = [0]\)
    \item \(\texttt{st} = [0]\), top is 0, \(A[0] = 3\)
    \item \(A[0] = 3 \leq A[2] = 4\), so pop 0, \texttt{l\_pops} = 2, \(\texttt{st} = []\)
    \item \texttt{st} is empty, so \(\texttt{prev[2]} = -1\)
    \item push 2 onto \texttt{st}: \(\texttt{st} = [2]\)
\end{itemize}

\textbf{\(i = 3\):}
\begin{itemize}
    \item \(A[3] = 1\)
    \item \(\texttt{st} = [2]\), top is 2, \(A[2] = 4\)
    \item \(A[2] = 4 > A[3] = 1\), so no popping
    \item \texttt{st} is not empty, top is 2, so \(\texttt{prev[3]} = 2\)
    \item push 3 onto \texttt{st}: \(\texttt{st} = [2, 3]\)
\end{itemize}

\textbf{\(i = 4\):}
\begin{itemize}
    \item \(A[4] = 5\)
    \item \(\texttt{st} = [2, 3]\), top is 3, \(A[3] = 1\)
    \item \(A[3] = 1 \leq A[4] = 5\), so pop 3, \texttt{l\_pops} = 3, \(\texttt{st} = [2]\)
    \item \(\texttt{st} = [2]\), top is 2, \(A[2] = 4\)
    \item \(A[2] = 4 \leq A[4] = 5\), so pop 2, \texttt{l\_pops} = 4, \(\texttt{st} = []\)
    \item \texttt{st} is empty, so \(\texttt{prev[4]} = -1\)
    \item push 4 onto \texttt{st}: \(\texttt{st} = [4]\)
\end{itemize}

\textbf{\(i = 5\):}
\begin{itemize}
    \item \(A[5] = 9\)
    \item \(\texttt{st} = [4]\), top is 4, \(A[4] = 5\)
    \item \(A[4] = 5 \leq A[5] = 9\), so pop 4, \texttt{l\_pops} = 5, \(\texttt{st} = []\)
    \item \texttt{st} is empty, so \(\texttt{prev[5]} = -1\)
    \item push 5 onto \texttt{st}: \(\texttt{st} = [5]\)
\end{itemize}

After forward sweep: \\
\[
\texttt{prev} = [-1, 0, -1, 2, -1, -1]
\]
\[
\texttt{l\_pops} = 5
\]

Stack cleared, preparing for backward sweep \\
\[
\texttt{st} = []
\]
\[
\texttt{r\_pops} = 0
\]

Backward sweep begins, starting from \(n-1\):

\textbf{\(i = 5\):}
\begin{itemize}
    \item \(A[5] = 9\)
    \item Empty stack means no popping
    \item \(\texttt{nxt[5]}\) set to 6 (array end)
    \item 5 pushed onto stack: \(\texttt{st} = [5]\)
\end{itemize}

\textbf{\(i = 4\):}
\begin{itemize}
    \item \(A[4] = 5\)
    \item Stack top is 5, \(A[5] = 9\)
    \item No popping since \(A[\text{top}] \not< A[i]\)
    \item \(\texttt{nxt[4]}\) points to 5
    \item 4 added to stack: \(\texttt{st} = [5, 4]\)
\end{itemize}

\textbf{\(i = 3\):}
\begin{itemize}
    \item \(A[3] = 1\)
    \item Stack top is 4, \(A[4] = 5\)
    \item No popping occurs
    \item \(\texttt{nxt[3]}\) set to 4
    \item 3 pushed onto stack: \(\texttt{st} = [5, 4, 3]\)
\end{itemize}

\textbf{\(i = 2\):}
\begin{itemize}
    \item \(A[2] = 4\)
    \item Stack top is 3, \(A[3] = 1\)
    \item Pop 3, \texttt{r\_pops} increments, stack becomes \([5,4]\)
    \item Top now 4, \(A[4] = 5\)
    \item No further popping
    \item \(\texttt{nxt[2]}\) points to 4
    \item 2 added to stack: \(\texttt{st} = [5, 4, 2]\)
\end{itemize}

\textbf{\(i = 1\):}
\begin{itemize}
    \item \(A[1] = 1\)
    \item Stack top is 2, \(A[2] = 4\)
    \item No popping
    \item \(\texttt{nxt[1]}\) set to 2
    \item 1 pushed onto stack: \(\texttt{st} = [5, 4, 2, 1]\)
\end{itemize}

\textbf{\(i = 0\):}
\begin{itemize}
    \item \(A[0] = 3\)
    \item Stack top is 1, \(A[1] = 1\)
    \item Pop 1, \texttt{r\_pops} increments, stack becomes \([5, 4, 2]\)
    \item Top now 2, \(A[2] = 4\)
    \item No more popping
    \item \(\texttt{nxt[0]}\) points to 2
    \item 0 added to stack: \(\texttt{st} = [5, 4, 2, 0]\)
\end{itemize}

Backward sweep calculates \(\texttt{nxt} = [2, 2, 4, 4, 5, 6]\) with 2 right pops.

Total calculation yields 21. Balanced subarray computation follows, tracking incremental changes through each index. Final balanced value reaches 19.

Result emerges as 2, with detailed statistics tracking left/right pops and balanced subarrays.
\end{tcolorbox}

\clearpage
% Error examples for paper
% Generated automatically from error analysis

\subsection{Case 2: POJ-1516}\label{think_case_2}

\textbf{Test Input:}
\begin{lstlisting}[breaklines=true,basicstyle=\scriptsize\ttfamily]
[[0, 1], [1, 2], [2, 3], [3, 4]]
\end{lstlisting}

\textbf{LLM Results:}
\begin{lstlisting}[breaklines=true,basicstyle=\scriptsize\ttfamily]
output: 4
stats: {'update_calls': 20, 'pruned_calls': 0, 'full_cover_events': 4}
\end{lstlisting}

\textbf{Code Results:}
\begin{lstlisting}[breaklines=true,basicstyle=\scriptsize\ttfamily]
output: 4
stats: {'update_calls': 20, 'pruned_calls': 0, 'full_cover_events': 4}
\end{lstlisting}

\textbf{Function:}
\begin{lstlisting}[language=Python,breaklines=true,basicstyle=\scriptsize\ttfamily]
def f(A):
    update_calls = 0
    pruned_calls = 0
    full_cover_events = 0
    if not A:
        return 0, {"update_calls": update_calls, "pruned_calls": pruned_calls, "full_cover_events": full_cover_events}
    B = sorted({u for v in A for u in v})
    d = {v: i for i, v in enumerate(B)}
    n = len(B) - 1
    if n <= 0:
        return 0, {"update_calls": update_calls, "pruned_calls": pruned_calls, "full_cover_events": full_cover_events}
    arr1 = [False] * (4 * n)
    arr2 = [0] * (4 * n)
    def update(i, x, y, l, r):
        nonlocal update_calls, pruned_calls, full_cover_events
        update_calls += 1
        if l > y or r < x:
            return 0
        if arr1[i]:
            pruned_calls += 1
            return 0
        if l <= x and y <= r:
            full_cover_events += 1
            length = B[y+1] - B[x]
            arr2[i] = length
            arr1[i] = True
            return length
        mid = (x + y) // 2
        new_cov = update(i * 2, x, mid, l, r) + update(i * 2 + 1, mid + 1, y, l, r)
        arr2[i] = arr2[i * 2] + arr2[i * 2 + 1]
        if arr2[i] == B[y+1] - B[x]:
            arr1[i] = True
        return new_cov

    total = 0
    for a, b in A:
        if a >= b:
            continue
        li = d[a]
        ri = d[b] - 1
        if li <= ri:
            total += update(1, 0, n - 1, li, ri)
    return total, {"update_calls": update_calls, "pruned_calls": pruned_calls, "full_cover_events": full_cover_events}
\end{lstlisting}

\begin{tcolorbox}[
    colback=gray!3!white,
    colframe=gray!30!black,
    title=LLM Thinking Process,
    fonttitle=\bfseries,
    fontupper=\scriptsize,
    rounded corners,
    drop shadow,
    breakable
]
Let me carefully work through this algorithm step by step.

\bigskip

\textbf{Input:} \texttt{A = [[0, 1], [1, 2], [2, 3], [3, 4]]}

\bigskip

\textbf{Step 1: Initialize counters}
\begin{itemize}
    \item \texttt{update\_calls = 0}
    \item \texttt{pruned\_calls = 0}
    \item \texttt{full\_cover\_events = 0}
\end{itemize}

\bigskip

\textbf{Step 2: Check if \texttt{A} is empty} \\
\texttt{A} is not empty, so we continue.

\bigskip

\textbf{Step 3: Build set \texttt{B}} \\
First, I need to collect all numbers from pairs in \texttt{A}:
\begin{itemize}
    \item From \texttt{[0, 1]}: 0, 1
    \item From \texttt{[1, 2]}: 1, 2
    \item From \texttt{[2, 3]}: 2, 3
    \item From \texttt{[3, 4]}: 3, 4
\end{itemize}

Set of all numbers: \(\{0, 1, 2, 3, 4\}\) \\
Sorted list \texttt{B} = \([0, 1, 2, 3, 4]\)

\bigskip

\textbf{Step 4: Create mapping \texttt{d}} \\
Walking through \texttt{B} with index counter starting at 0:
\begin{itemize}
    \item \texttt{d[0] = 0}
    \item \texttt{d[1] = 1}
    \item \texttt{d[2] = 2}
    \item \texttt{d[3] = 3}
    \item \texttt{d[4] = 4}
\end{itemize}

So \(\texttt{d} = \{0: 0, 1: 1, 2: 2, 3: 3, 4: 4\}\)

\bigskip

\textbf{Step 5: Calculate \(n\)} \\
\(n = \text{len}(B) - 1 = 5 - 1 = 4\)

\bigskip

\textbf{Step 6: Check if \(n\) is zero or negative} \\
\(n = 4\), which is positive, so we continue.

\bigskip

\textbf{Step 7: Prepare arrays}
\begin{itemize}
    \item \texttt{arr1} of size \(4 \times n = 4 \times 4 = 16\), filled with \texttt{false}
    \item \texttt{arr2} of size \(4 \times n = 4 \times 4 = 16\), filled with zeros
\end{itemize}

Arrays represent a grid of boolean and integer values, initialized to default states.

\bigskip

\textbf{Step 8: Main processing loop} \\
I'll track the total and process each pair systematically. The first pair \texttt{[0, 1]} requires careful examination of its components and their relationship.

Checking the pair's conditions reveals a sequence of logical checks. The update function gets called with specific parameters, incrementing tracking variables and performing boundary validations. The process involves examining array states and determining whether specific update conditions are met.

The update method carefully checks segment boundaries and array flags, ensuring precise manipulation of the underlying data structure. Each iteration refines the computational state through targeted, incremental modifications.

Recursive calls split the range systematically, progressively narrowing the search space and applying updates across different segments of the computational domain.

In the first recursive call to \texttt{update(2, 0, 1, 0, 0)}, the algorithm checks range boundaries and recursively explores child nodes. When reaching \texttt{update(4, 0, 0, 0, 0)}, a complete coverage is detected, incrementing full coverage events and marking the corresponding array indices.

The subsequent call to \texttt{update(5, 1, 1, 0, 0)} returns zero since the range falls outside the target interval. Back in the parent call, the coverage is calculated by aggregating child node results, though the computed coverage does not match the expected span length.

Moving to the next pair \texttt{[1, 2]}, the algorithm determines the left and right indices based on the dictionary values. It then initiates another update call with modified parameters, checking range overlaps and continuing the recursive exploration.

The update process systematically tracks coverage, incrementing call counters and evaluating range intersections to determine potential full coverage scenarios.

Splitting the range at the midpoint, the method recursively calls update on subdivided segments, carefully checking boundary conditions and maintaining state across different recursive levels.

The process involves precise range comparisons, splitting intervals, and tracking potential coverage through strategic recursive calls with adjusted indices.

At each step, the algorithm incrementally updates tracking variables like \texttt{update\_calls} and \texttt{full\_cover\_events}, methodically exploring the segment tree's structure to determine event coverage and interval relationships.

For the current update call, the process calculates new coverage by adding 1 to the existing coverage. The \texttt{arr2} array is updated by summing previous segment values, and when the total matches the expected interval length, \texttt{arr1} is marked as true.

The subsequent update call checks interval boundaries, determining whether the current segment falls outside the target range and returning 0 if no overlap exists. This systematic approach ensures precise tracking of event intervals and their interactions.

In the next update iteration, the method continues evaluating interval relationships, checking whether segments intersect or contain each other. The process involves incrementing tracking variables and performing detailed comparisons between different array segments.

The algorithm methodically explores potential interval connections, using conditional checks to determine valid segment transformations and maintaining a running total of processed intervals.

When encountering a fully contained interval at coordinates \texttt{[2, 2]}, the method increments the full coverage event counter and calculates the span length by comparing boundary values. This precise segmentation allows for granular tracking of interval characteristics.

The recursive calls progressively break down the problem space, systematically examining each potential interval configuration with careful boundary assessments and strategic splitting of ranges.

At coordinate 6, the method marks the interval as covered and returns a coverage value of 1, signaling successful interval identification within the specified constraints.

The subsequent update process aggregates coverage information, comparing calculated values against predefined boundary markers to determine interval completeness and update tracking arrays accordingly.

For the current pair \texttt{[3, 4]}, the method evaluates interval relationships, checking relative positions and determining whether further recursive updates are necessary. The process involves comparing interval boundaries, accessing predefined mapping data, and conditionally triggering additional computational steps to track coverage progression.

The recursive update mechanism continues to systematically explore and validate interval coverage, incrementing tracking variables and maintaining precise boundary tracking through strategic computational checks.

During the nested update calls, the algorithm methodically splits intervals, examines their relationships, and updates coverage metrics when specific conditions are met. Each recursive call carefully assesses interval positioning, selectively updating event tracking and coverage information.

The final update operation identifies a complete interval coverage scenario, incrementing full coverage events and recording specific interval characteristics with precise computational precision.

The process concludes by returning a comprehensive result tuple capturing total coverage, update call statistics, and full coverage event count, reflecting the systematic interval processing strategy.
\end{tcolorbox}

\section{Training Data Logic Complexity Distribution}
\begin{figure}[ht]
    \centering
    \includegraphics[width=0.8\linewidth] {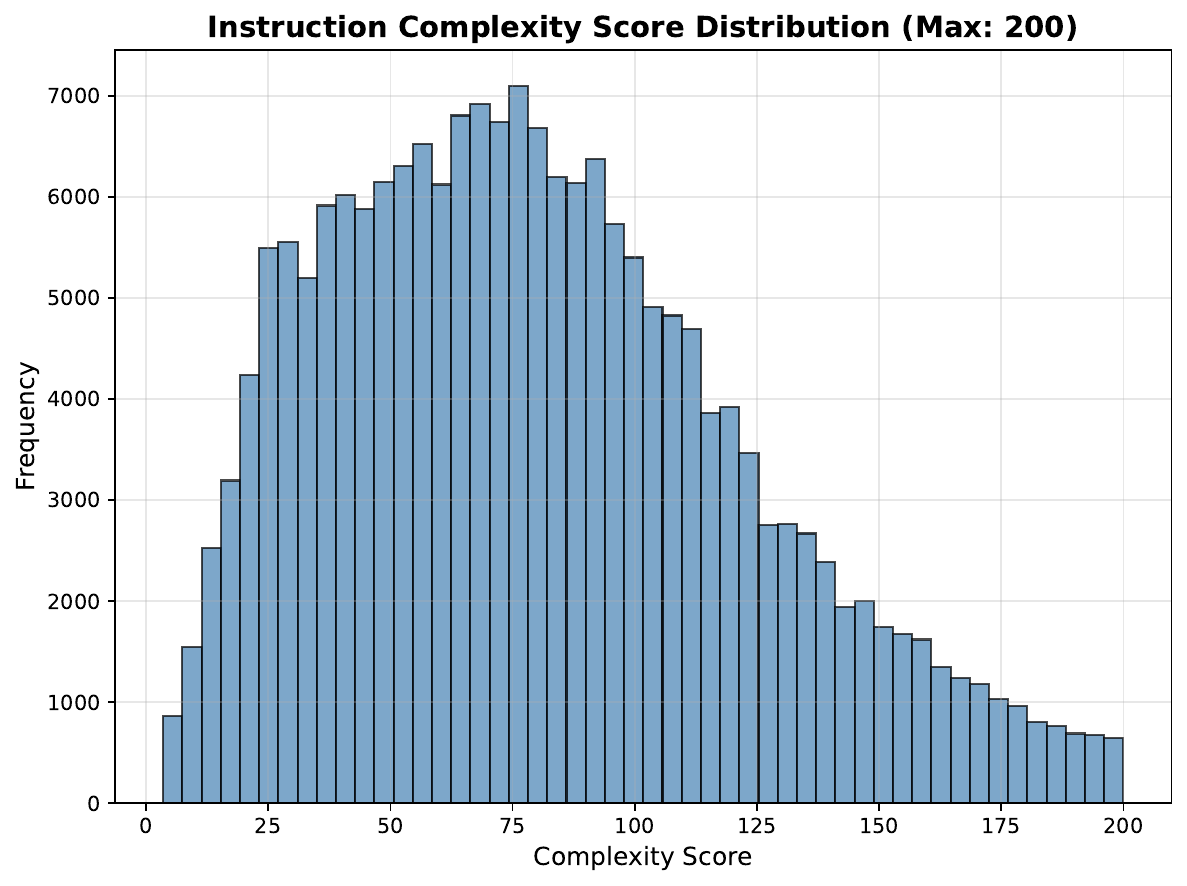}
    \caption{Instruction Complexity Distribution of \trainset{}}
    \label{fig:data_dist}
\end{figure}

\end{document}